\documentclass[journal]{IEEEtran}

\usepackage{cite,graphicx,amsmath,amssymb}
\usepackage{comment}
\usepackage{graphicx}
\usepackage{amsfonts,amssymb}
\usepackage{xcolor}
\usepackage{multirow}
\usepackage{url}
\usepackage{enumerate}
\usepackage{enumitem}
\usepackage[
top    = 0.67in,
bottom = 0.94in,
left   = 0.63 in,
right  = 0.63 in]{geometry}
\usepackage{xspace}
\usepackage{diagbox}

\usepackage{algpseudocode}
\usepackage{algorithm}
\usepackage{booktabs}
\usepackage{caption}
\usepackage{subfig} 

\usepackage[colorlinks=true, linkcolor=blue, urlcolor=blue, citecolor=blue]{hyperref}

\title{Causal-Aware Intelligent QoE Optimization for VR Interaction with Adaptive Keyframe Extraction}

\author{Ziru Zhang,~\IEEEmembership{Student Member,~IEEE},
        Jiadong~Yu,~\IEEEmembership{Member,~IEEE},
        and Danny H.K.~Tsang,~\IEEEmembership{Life Fellow,~IEEE}
\IEEEcompsocitemizethanks{\IEEEcompsocthanksitem Ziru Zhang and Jiadong~Yu are with The Internet of Things Thrust, The Hong Kong University of Science and Technology (Guangzhou),  Guangzhou, China (email: zzhang758@connect.hkust-gz.edu.cn; jiadongyu@hkust-gz.edu.cn). 

Danny H.K.~Tsang is with The Internet of Things Thrust, The Hong Kong University of Science and Technology (Guangzhou), Guangzhou, China, and also with The Department of Electronic and Computer Engineering, The Hong Kong University of Science and Technology, Hong Kong (email: eetsang@ust.hk). \protect\\}}

\begin{document}
\maketitle
 \thispagestyle{empty}
 \pagestyle{empty}

\begin{abstract}
The optimization of quality of experience (QoE) in multi-user virtual reality (VR) interactions demands a delicate balance between ultra-low latency, high-fidelity motion synchronization, and equitable resource allocation. While adaptive keyframe extraction mitigates transmission overhead, existing approaches often overlook the causal relationships among allocated bandwidth, CPU frequency, and user perception, limiting QoE gains. This paper proposes an intelligent framework to maximize QoE by integrating adaptive keyframe extraction with causal-aware reinforcement learning (RL). First, a novel QoE metric is formulated using the Weber-Fechner Law, combining perceptual sensitivity, attention-driven priorities, and motion reconstruction accuracy. The QoE optimization problem is then modeled as a mixed integer programming (MIP) task, jointly optimizing keyframe ratios, bandwidth, and computational resources under horizon-fairness constraints. We propose Partial State Causal Deep Deterministic Policy Gradient (PS-CDDPG), which integrates the Deep Deterministic Policy Gradient (DDPG) method with causal influence detection. By leveraging causal information regarding how QoE is influenced and determined by various actions, we explore actions guided by weights calculated from causal inference (CI), which in turn improves training efficiency. Experiments conducted with the CMU Motion Capture Database demonstrate that our framework significantly reduces interactive latency, enhances QoE, and maintains fairness, achieving superior performance compared to benchmark methods.

\end{abstract}
\textbf{Keywords:} Quality of Experience, Virtual Reality, Wireless Communication, Deep Reinforcement Learning, Causal Reinforcement Learning.
\section{Introduction}
The virtual reality (VR) industry has experienced substantial growth in recent years, leading to groundbreaking applications across various domains. As a technology centered on human experience and immersion, VR has prioritized real-time multiplayer interactions as a pivotal application area. Innovations in 3D and multimedia technologies have enabled VR to surpass the limitations of traditional video-based communication, offering a broader range of immersive experiences. However, the advancement of VR technology faces significant constraints due to its requirements for ultra-high definition and sensitivity to latency. Furthermore, the transmission of human movements in three-dimensional environments not only enhances the sophistication of VR features but also increases the demand for data flow resources. To address these challenges, researchers are exploring the implementation of attention mechanisms within VR systems. These mechanisms are designed to optimize data communication processes by focusing on the user's field of view (FoV), ultimately reducing data volume and latency.

Virtual scene rendering relies on 3D models and motion sequences, which serve as a crucial foundation for animation, VR, and the Metaverse. Through technologies such as cameras or inertial measurement unit (IMU) sensors, human motion data is captured and essential for synchronizing user movements from physical environment to virtual scene. However, despite their high precision in capturing action details, these motion sequences impose significant demands on communication and synchronization. To address these issues, researchers are exploring next-generation transmission technologies and leveraging the computational power of edge servers to improve quality of service (QoS). Furthermore, the reliability of reconstructing intermediate frames has led to the prominence of keyframe extraction and motion reconstruction as vital techniques \cite{keyframe_1} \cite{keyframe_2}.

While traditional QoS focuses on technical standards such as error rate and latency, quality of experience (QoE) serves as a user-centered metric that assesses overall user satisfaction and perception. By emphasizing how well a service meets user expectations, QoE facilitates the rational allocation and utilization of resources. This is particularly crucial in achieving immersive experiences in VR, especially in resource-constrained environments \cite{QoE_VR_1}. However, the adaptation of multimedia approaches complicates the proposal of a QoE function due to the randomness of user behavior, perceptual thresholds, and the need for fairness and stability in perceived services. Additionally, the level of user attention to different objects also determines the priority for resource allocation, making the enhancement of QoE in VR interactions even more challenging.

The expansion of artificial intelligence is revolutionizing various fields, particularly in intelligent decision-making. Leveraging the real-time capabilities of smart agents is essential for optimizing user experiences during multi-user interactions within dynamic network environments. Among machine learning algorithms, reinforcement learning (RL) is notable for its ability to address complex dynamic systems. By interacting with the environment, RL agents can learn with less reliance on extensive training datasets. However, a major challenge in RL is training efficiency, which often involves extensive iterative processes to adapt to new environments. Causal RL (CRL), which adapts causal inference (CI) to empower the RL training process, has also emerged as a new research trend to enhance both efficiency and interpretability. By detecting the impact of actions on the current state, the agent can identify training strategies, thereby reducing randomness during training \cite{CRL_0}.

Inspired by the above facts, this paper proposes the QoE matrix in multi-user VR interactions leveraging the Weber-Fechner Law. The attention-based strategy and keyframe-based communication approach are designed to fully utilize the limited resources. To achieve this, the QoE maximization problem is formulated as a mixed-integer programming (MIP) problem, which jointly optimizes keyframe ratios, bandwidth, and computational resources while ensuring fairness among users. A novel decision-making algorithm is proposed based on the Deep Deterministic Policy Gradient (DDPG) model and causal influence detection. The causal action influence (CAI) score is utilized to quantify the causal information on how the given state is influenced by actions. Following this, a noise-based active exploration scheme is proposed, which selects actions based on weights derived from the CAI scores of different candidate actions. The major contributions of this paper can be summarized as follows:

\begin{itemize}
    \item We model a multi-user VR interaction system under a sub-6 GHz communication environment. The movements of users are captured and transmitted using a keyframe extraction and motion reconstruction approach. We then propose a QoE function that integrates the Weber-Fechner Law, attention strategies, and keyframe ratios.
    \item We formulate an optimization problem for QoE enhancement by jointly optimizing the keyframe ratio and the allocation of bandwidth and CPU frequency. This problem is classified as a MIP problem, which trade-offs between sequence accuracy and communication latency to achieve both near-optimal QoE and horizon-fairness QoE (hfQoE).
    \item We propose a Partial State - Causal DDPG (PS-CDDPG) algorithm to intelligently address the problem. The state variables are divided into action-relevant and action-irrelevant categories. The causal influence score is inferred using a deep neural network (DNN), which serves as the inference model for CI. The RL agent is then trained with noise-based active exploration, which leverages causal information to enhance learning effectiveness.
    \item The framework is evaluated through extensive experiments using a human motion dataset. The numerical results demonstrate that the proposed framework significantly improves training efficiency. Additionally, QoE is enhanced while maintaining fairness among users and outperforming baseline methods.
\end{itemize}

The remainder of this paper is organized as follows. In Section~\ref{related_work}, we summarize related work on QoE in VR and intelligent algorithms. Next, we elaborate the system model and problem formulation in Section~\ref{system_model}. In Section~\ref{algorithm}, we present the proposed PS-CDDPG algorithm. Then, we demonstrate the evaluation process and present the numerical results in Section~\ref{evaluation}. Finally, we conclude the paper in Section~\ref{conclusion} and discuss future work.
\section{Related Work}
\label{related_work}
VR technologies utilize head-mounted displays (HMDs) to enhance both the FoV and resolution, enabling users to perceive expansive and detailed virtual environments. Within these environments, virtual characters are rendered using advanced engines that incorporate models and motion sequences. To optimize QoE in VR interactions, it is crucial to consider unique multimedia mechanisms and data flows during the resource allocation process. Furthermore, the ability to navigate complex communication environments and support intelligent decision-making is crucial for enhancing user engagement.

\subsection{Keyframe Extraction in VR}
Valued for its application potential, VR has been widely used in areas such as VR gaming\cite{VR_game}, virtual meetings\cite{VR_meeting}, and the Metaverse\cite{Metaverse}. By precisely capturing users' movements, motion capture enables the creation of digital representations that closely mimic users' actions. To enhance the precision of this process, high-precision IMU sensors and advanced computer vision techniques are leveraged, allowing motion sequences to be accurately recorded and saved in Biovision Hierarchy (BVH) format. However, the fundamental approach to synchronizing movements in VR involves transmitting a substantial amount of motion data in a short time, which inevitably leads to noticeable delays. Consequently, keyframe extraction and motion reconstruction techniques can be employed as effective methods to minimize redundant data and reduce latency.

Keyframe extraction techniques, which identify a set of representative frames, are commonly used for various applications, including video classification \cite{keyframe_video_classification}, summarization \cite{keyframe_video_summarization}, and reconstruction \cite{keyframe_video_reconstruction}. However, due to the high uncertainty between pixels, the robustness and efficiency of using keyframes to transmit video are insufficient in practice. In contrast, keyframe-based communication of motion sequences has great potential, which arises from the continuous nature of human movements and joint coordinates. For instance, the work in \cite{keyframe_1} introduced the learning-based sphere nonlinear interpolation (Slerp) method for motion reconstruction using keyframes. In this approach, all coordinates are transformed into a spherical coordinate system, and spherical interpolation is used to generate in-between motions. Xia et al. \cite{keyframe_2} proposed the joint kernel sparse representation model for keyframe extraction, which effectively models both the sparsity and the Riemannian manifold structure of human motion to decrease the unreasonable distribution and redundancy of extracted keyframes. Furthermore, the work in \cite{keyframe_3} integrated subspace learning and self-expression approaches with sphere interpolation, facilitating the learning of explicit features of motion data and further enhancing performance. Moreover, deep Q-learning and long-short-term memory (LSTM) were utilized for graph-based human motion data \cite{keyframe_4}. The normalized reconstruction error of sphere interpolation is designed as a reward function to train the RL model.

Numerous studies have also explored the application during the communication process \cite{keyframe_communication_1}, \cite{keyframe_communication_2}, \cite{keyframe_communication_3}, \cite{keyframe_communication_4}, \cite{keyframe_communication_5}. The work in \cite{keyframe_communication_1} proposed a keyframe-based coding system for transmission over differentiated services networks, where a higher QoS priority is assigned to keyframe packets. In the work presented in \cite{keyframe_communication_2}, an edge-assisted communication environment was explored, focusing on keyframe selection to minimize transmission delay while maintaining data accuracy. A novel hierarchical deep reinforcement learning (HDRL) framework was employed to jointly determine the routing path and keyframes. Furthermore, an end-to-end video compression scheme for low-latency scenarios was proposed in \cite{keyframe_communication_3}, incorporating motion prediction, compensation, and reconstruction techniques. Leveraging the capabilities of generative adversarial networks (GANs), the work in \cite{keyframe_communication_4} presented a novel framework for video compression and anomaly detection, utilizing inter-frame prediction and frame reconstruction. In \cite{keyframe_communication_5}, a resource-aware DRL framework was designed to optimize transmission in mobile augmented reality (AR), balancing the trade-off between processing frame rate and accuracy.

The works mentioned above demonstrate that most schemes of keyframe-based communication are designed for video transmission. The keyframe extraction and motion reconstruction approaches also show high reliability and efficiency. However, the balance between the number of keyframes and the reconstruction of motion sequences in VR interactions still requires further discussion. In addition, state-of-the-art keyframe extraction methods indicate that reconstruction accuracy is primarily influenced by the keyframe ratio \cite{keyframe_2} rather than by the extraction methods themselves. Although advanced extraction algorithms can improve accuracy, their complex processes may also introduce additional computational burdens that impact transmission latency. Additionally, a higher keyframe rate is likely to result in greater accuracy. In our work, we focus on jointly optimizing the ratio of keyframes during transmission and the allocation of various resources to achieve an enhanced experience.
\vspace{-.2cm}

\subsection{Intelligent QoE Optimization}
QoE in VR applications encompasses various aspects such as visual fidelity, latency, frames per second (FPS), and interactivity, all of which significantly impact user satisfaction and engagement. To address these diverse factors, numerous works have defined customized QoE matrices for various VR application scenarios, including Metaverse \cite{QoE_1_metaverse}, VR content streaming \cite{QoE_2_360}, \cite{QoE_VR_1}, and VR games \cite{QoE_5_game}. For instance, the work \cite{QoE_1_metaverse} explored attention-aware Metaverse services within a multiple-input multiple-output (MIMO) system. A new QoE matrix was derived by combining key performance indicators with the Weber–Fechner Law \cite{wb_1}. Similarly, utilizing the viewpoint and FoV of the users, \cite{QoE_2_360} reduced the transmission redundancy for 360-degree video streaming to maximize QoE. Moreover, digital twin techniques were applied to enhance MEC-enabled VR content streaming in \cite{QoE_VR_1}, where attention mechanisms and latency under the sub-6 GHz communication model were incorporated into the QoE matrix. In the context of VR games, the study \cite{QoE_5_game} defined QoE as the importance weight assigned to various objects and their corresponding rendering levels. 

Although the QoE matrix reasonably integrates standards from different aspects, the unavoidable complexity poses challenges in solving optimization problems. Consequently, a collection of artificial intelligence (AI) algorithms has been implemented to provide intelligent decisions for QoE maximization \cite{QoE_2_360}, \cite{QoE_VR_1}, \cite{QoE_transformer}, \cite{QoE_diffusion}. Additionally, various techniques have been combined with RL algorithms to enhance performance, including continuous learning \cite{QoE_VR_1}, transformers \cite{QoE_transformer}, and diffusion models \cite{QoE_diffusion}. By interacting with the environment, the agent can be trained automatically, enabling it to handle complex and dynamic application scenarios. However, despite these advancements, the large number of interaction iterations and the slow learning speed remain major challenges that limit its development.

CRL, as an extension of RL techniques, primarily aims to leverage causal information within the Markov decision process (MDP) to guide model training \cite{CRL_survey}. In this context, numerous studies have explored the effective deployment of CI to support RL training \cite{CRL_0}, \cite{CRL_1}, \cite{CRL_2}, \cite{CRL_3}, \cite{CRL_4}. Among these studies, the causal influence detection method proposed by \cite{CRL_0} offers a robust solution for enhancing training efficiency. Unlike traditional RL algorithms that focus primarily on action decision-making, this approach emphasizes the analysis of how different actions influence a given state. Specifically, if the influence is significant, it indicates that the next state is strongly affected by the action taken. This suggesting that the agent has a strong control over the environment, thereby making the data more valuable for training. Conversely, if the influence is weak, it implies that the current state is less controllable and lacks training value. This limitation hinders the agent's ability to learn useful experiences from the corresponding actions. Consequently, the obtained influence values can effectively guide the training process in three distinct ways: active exploration, reward bonus, and experience replay.

Most works on QoE customization only consider the transmission of rendered panoramic videos or images \cite{QoE_1_metaverse},\cite{QoE_2_360},\cite{QoE_VR_1}. However, in multi-user interaction scenarios, it is crucial to transmit and synchronize only the necessary data, such as user actions, game states, and text \cite{synchronization}. Furthermore, inspired by the attention-aware resource allocation approaches in \cite{Metaverse} and \cite{QoE_VR_1} for tiled 360-degree video transmission, we propose to integrate the attention mechanism with human motion sequences through keyframe extraction and motion reconstruction. By allocating a higher keyframe ratio to objects that attract more attention, we can effectively manage data volume and optimize bandwidth usage. Moreover, while the case studies discussed in \cite{CRL_0} mainly focus on the performance of various fundamental environments, our objective is to assess the potential of a causal influence detection-empowered RL algorithm for QoE maximization. This approach introduces increased complexity in both decision-making processes and environmental feedback mechanisms.
\vspace{-.2cm}

\section{System Model}
\label{system_model}
\begin{figure*}
    \centering
    \includegraphics[width=1\linewidth]{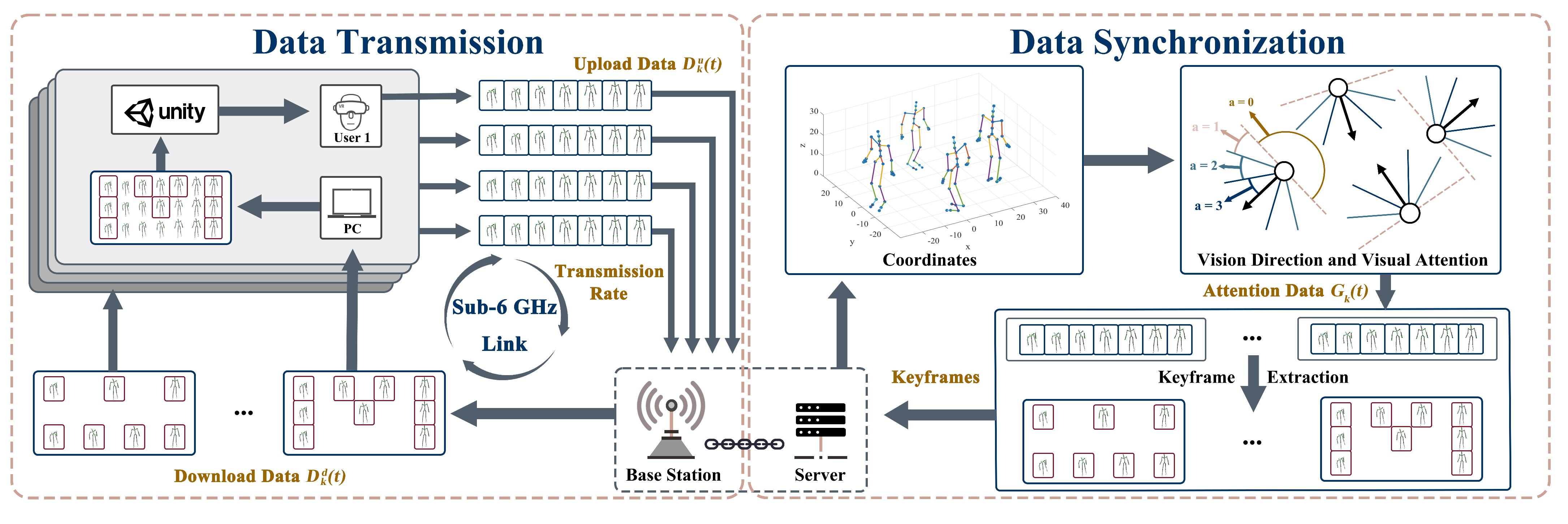}
    \caption{The framework of the attention-based keyframe extraction for VR interaction.}
    \label{Framework_old}
    \vspace{-.4cm}
\end{figure*}

In this section, we provide a detailed explanation of the proposed multi-user interaction system model of VR, focusing on the challenge of optimizing QoE. As illustrated in Fig. \ref{Framework_old}, the system facilitates real-time interactions among multiple users within a shared environment. A central server, equipped with an access point that employs Sub-6 GHz technology, is tasked with gathering and synchronizing motion data from users. This data, captured via motion capture devices on the user side, is stored in BVH format. Subsequently, the necessary motion sequences are sent back to the users, where they are processed and rendered using the local computational resources of each user. This entire workflow underscores our keyframe-based methodology and integrates user attention into the communication framework.

\begin{table}[ht]

\renewcommand{\tablename}{TABLE}
\caption{List of the Important Notations}   

\centering
\begin{tabular}{c|c}
\toprule
\textbf{Notations} & \textbf{Definitions}  \\
\midrule  
$a$ & Attention level\\
$k$, $K$ & User index and total number of users\\
$t$, $\Delta t$ & Time slot index and the duration of a time slot\\
$\xi$ & Frames per second of VR\\
$\delta$ & The file size for each frame of a single character\\
$N_{k,a}(t)$ & Characters of the $k^{th}$ user with attention level $a$\\
$F_{k,a}(t)$ & Keyframes transmitted per second for each character\\
$W^u_k(t)$& Upload data size of the $k^{th}$ user\\
$W^d_k(t)$& Download data size of the $k^{th}$ user\\
$c_e$ & CPU cycles for processing a bit for keyframe extraction\\
$c^1_r$ & CPU cycles for processing a bit for reconstruction \\
$c^2_r$ & CPU cycles for processing a bit for rendering\\
$\omega$ & Compression ratio for motion data transmission\\
$f^e_k(t)$ & Allocated CPU frequencies of server for the $k^{th}$ user\\
$f^r_k(t)$ & CPU frequencies of the VR device for the $k^{th}$ user\\
$ f_{\text{max}}$ & Total CPU frequencies of server\\
$T^u_k(t)$, $T^d_k(t)$ & Time latency of uploading and downloading data\\ 
$T^e_k(t)$ & Time latency of keyframe extraction\\
$T^r_k(t)$ & Time latency of reconstruction and rendering\\
$T_k(t)$, $T_{max}$ & Total latency of the $k^{th}$ user and time threshold\\
$R_k(t)$ & Transmission rate of the $k^{th}$ user\\
$b_k(t)$, $b_{\text{max}}$ & Subchannel bandwidth and total bandwidth\\
$P_k(t)$, $\sigma_k^2$ & Transmitter power and noise power\\
$g_k(t)$ & Channel gain\\
$\beta_k(t)$ & Large-scale fading coefficient\\
$h_k(t)$ & Small-scale fading coefficient\\
$d_0$ & Reference distance\\
$X_\sigma$ & Shadow fading with a standard deviation of $\sigma$\\
$\mathrm{QoE}_k(t)$ & QoE of the $k^{th}$ user at time $t$\\
$\mathrm{hfQoE}(t)$ & Horizon-fairness QoE\\
$H(t)$, $L(t)$ & Highest and lowest QoE\\
$\mathrm{QoE}_{\text{th}}$ & Threshold of QoE\\
$\mathrm{hfQoE}_{\text{th}}$ & Threshold of Horizon-fairness QoE\\
\bottomrule  
\end{tabular}
\label{notations}
\vspace{-.6cm}
\end{table}

\subsection{Visual Attention}
Due to the unique spherical nature of the virtual environment and the rendering process in VR systems, the same 3D motion sequence is perceived differently by each user, depending on their orientation of the head-mounted display (HMD). Given that each individual possesses a distinct FoV, it is essential to optimize resource utilization during both transmission and rendering. Leveraging the hierarchical nature of human vision and the limited FoV of current HMD devices, dynamically modifying resource allocation decisions according to attention levels can significantly minimize unnecessary data transmission. Consequently, we categorize horizontal visual attention in human vision into four distinct levels, denoted as $(a \in \alpha= {0, 1, 2, 3})$:
\begin{itemize}
    \item $a = 3$: Central vision (0°–30°).
    \item $a = 2$: Peripheral binocular vision (30°–60°).
    \item $a = 1$: Monocular vision (60°–90°).
    \item $a = 0$: Blind spot (90°–180°).
\end{itemize}

The central vision shares the highest level of attention, while the blind spot remains invisible. This categorization allows us to optimize resource allocation by prioritizing regions that require greater attention while minimizing expenditure in areas that receive less focus.
\vspace{-.1cm}

\subsection{Keyframe Extraction}
In VR interaction systems, keyframes are vital for effectively depicting human motion. A keyframe refers to a specific frame within an animation sequence that indicates the beginning and ending points of a transition. When it comes to human movement, keyframes encapsulate the essential postures or actions, while the frames in between can be interpolated or reconstructed to ensure fluid motion between these keyframes. This strategy significantly decreases the volume of data that must be transmitted, as it eliminates the need to send every frame in the sequence, concentrating on the transmission of only the most important frames.

The file size for each frame with a single character is denoted as $\delta$. Given that the system functions at a FPS of $\xi$, the total amount of data uploaded and downloaded for the $k^{th}$ user during a specified time slot $t$ can be represented as:
\begin{align}
\left\{
	\begin{aligned}
	&W^u_k(t)=\xi \delta \Delta t,  \\
	&W^d_k(t)=\sum_{a\in \alpha} N_{k,a}(t) F_{k,a}(t) \delta \Delta t.
	\end{aligned}
\right.
\end{align}
In this equation, $\Delta t$ denotes the duration of a single time slot, $N_{k,a}(t)$ signifies the number of characters within the FoV of the $k^{th}$ user, characterized by an attention level $a$. Additionally, $F_{k,a}(t)$ represents the total number of keyframes transmitted for each character with attention level $a$ over one second.

The tasks of keyframe extraction (performed on the server side) and non-keyframe reconstruction (conducted on the user side) entail extra time expenditures. To quantify this, the CPU frequencies for the $k^{th}$ user on the server and the user device are represented as $f_k^e(t)$ and $f_k^r(t)$. Consequently, the time required for extracting keyframes on the server side can be expressed as follows:
\begin{equation}
    T^e_k(t)=\frac{c_e \sum_{a} N_{k,a}(t) F_{k,a}(t)\delta \Delta t}{f^e_k(t)},
\end{equation}
where $c_e$ represents the number of cycles required to process a single bit of data during the keyframe extraction process.

Upon receiving the keyframes, the user device reconstructs the intermediate frames and generates the complete scene. Previous research has investigated the application of interpolation techniques \cite{keyframe_2} and machine learning approaches \cite{keyframe_4} to improve the precision of motion reconstruction based on extracted keyframes, showing that satisfactory outcomes can be achieved. Consequently, we propose that the time required for these operations is:
\begin{equation}
    T^r_k(t)=\frac{c^1_r \sum_a N_{k,a}(t)(\xi-F_{k,a}(t))\delta\Delta t+c^2_r\xi\delta\Delta t}{f^r_k(t)},
\end{equation}
where $c_r^1$ and $c_r^2$ denote the computational coefficients associated with the reconstruction and rendering of a single bit of data, respectively.
\vspace{-.1cm}

\subsection{Sub-6 GHz Communication Model}
For the communication process, the $k^{th}$ user is linked to the base station through a sub-6 GHz connection. The theoretical transmission rate can be expressed as follows:
\begin{equation}
    R_k(t)=b_k(t)log_2\left(1+\frac{P_k(t) |g_k(t)|^2}{\sigma_k^2}\right),
\end{equation}
where $b_k(t)$ represents the subchannel bandwidth, $P_k(t)$ signifies the transmitter power, $g_k(t)$ indicates the channel gain, and $\sigma_k^2$ denotes the background noise power. The function $g_k(t)$ for the specified indoor environment follows the equation from \cite{sub6G_1}:
\begin{equation}
    g_k(t)= \sqrt{\beta_k(t)}h_k(t),
\end{equation}
where $\beta_k(t)$ denotes the large-scale fading coefficient, and $h_k(t)$ represents the small-scale fading coefficient, which follows the distribution $h_k(t)\sim \mathcal{CN}(0,1)$.

The large-scale fading coefficient $\beta_k(t)$, commonly referred as path loss, can be calculated using the free space reference distance path loss model of \cite{sub6G_2}:
\begin{equation}
    \beta_k(t)[dB]= PL(d_0)+10nlog_{10}\frac{d}{d_0}+X_\sigma, \quad d>d_0,
    \label{pl}
\end{equation}
where $d_0$ is the reference distance, $PL(d_0)$ is the free space path loss at $d_0$, $n$ is the path loss exponent that varies based on the environment, and $X_\sigma$ denotes shadow fading. Shadow fading is characterized by a zero-mean Gaussian distribution, with a standard deviation of $\sigma$ measured in dB.

Therefore, the communication latency, consisting of upload latency $T^u_k(t)$ and download latency $T^d_k(t)$ between the $k^{th}$ user and the base station, can be formulated as:
\begin{equation}
    T^u_k(t)+T^d_k(t)=\frac{D^u_k(t)+D^d_k(t)}{\omega R_k(t)},
\end{equation}
where $\omega$ represents the compression ratio for motion data.

    \vspace{-.2cm}
\subsection{QoE Definition}
The system model outlined above indicates that the QoE of VR interactions using keyframes and attention mechanisms is largely influenced by total latency $T_k(t)$, level of attention, and the number of keyframes. Inspired by the Weber-Fechner Law and related research \cite{wb_1}, we introduce a customized QoE evaluation function for users:
\begin{equation}
    \mathrm{QoE}_k(t)=\left( 1-\frac{T_k(t)}{T_{max}}\right)\sum_a\frac{aN_{k,a}(t)}{N_k(t)}ln\left(\frac{F_{k,a}(t)\Delta t}{2}\right).
\end{equation}
In this equation, $N_k(t)$ refers to the total number of characters present in the virtual scene, $T_{max}$ indicates the latency threshold, and the value $2$ represents the minimum required keyframes to be transmitted for each motion sequence.

The total latency experienced by the $k^{th}$ user can be determined using the following formula:
\begin{equation}
    T_k(t)=T_k^{u}(t)+ T^e_k(t)+T^d_k(t)+T^r_k(t).
\end{equation}
If $T_k(t)\leq T_{\mathrm{max}}$, it indicates that the requested FoV has been delivered and displayed successfully. In contrast, if $T_k(t) > T_{\mathrm{max}}$, it indicates a failure to deliver the requested FoV, resulting in a user QoE of 0. We then use $D_k(t)=\{T_k^{u}(t), T^e_k(t), T^d_k(t), T^r_k(t)\}$ to denote the delay matrix.

Furthermore, when maximizing overall QoE, it is vital to consider fairness among users to ensure that no user's experience is disproportionately lower than that of others. The horizon-fairness QoE can be expressed as follows \cite{hfQoE_1} \cite{hfQoE_2}:
\begin{equation}
    \mathrm{hfQoE}(t)=1-\frac{2\sigma^*}{H(t)-L(t)},
\end{equation}
where the standard deviation $\sigma^*$ is derived from the average QoE $avgQoE_k(t)$. $H(t) = \max\{QoE_k(t), k \in \mathcal{K}, t \in [0, t]\}$ and $L(t) = \min\{QoE_k(t), k \in \mathcal{K}, t \in [0, t]\}$ represent the highest and lowest QoE for $K$ users.

\subsection{Problem Formulation}
According to recent works \cite{keyframe_1} \cite{keyframe_4}, increasing the number of keyframes improves the accuracy of motion sequence reconstruction, particularly for intricate or rapid movements. When the keyframe ratio exceeds 10\%, the average Euclidean reconstruction error can be maintained within 1 degree \cite{keyframe_2}. In the proposed scheme, the server extracts keyframes during synchronization, and the intermediate frames are reconstructed before rendering. This approach effectively balances the accuracy of the reconstructed sequence and the volume of transmitted data, rendering it highly compatible with attention-aware techniques. By assigning a higher bandwidth and a greater number of keyframes to areas of greater visual importance (i.e., regions that draw more user attention within the FoV), we can improve overall QoE.

To enhance long-term QoE in a multi-user VR interaction environment, we developed a MIP optimization model, which aims to concurrently identify the optimal distribution of bandwidth $\textit{\textbf{B}} = { b_k(t) }$, CPU frequency $\textit{\textbf{f}} = { f_k^e(t) }$, and the number of keyframes $\textit{\textbf{F}} = { F_{k,a}(t) }$ for each user $k$ across the time slots:
\begin{subequations}
\begin{align}
(\mathcal{P}) \quad & \max_{B,f,F} \quad\sum_{t=0}^{\infty}\sum_{k=1}^K \mathrm{QoE}_k(t) \\
\text{s.t.:} \quad & \sum_{k} b_k(t) \leq b_{\text{max}}, \quad \forall t, \label{constraint1}\\
                    & \sum_{k} f^e_k(t) \leq f_{\text{max}}, \quad \forall t, \label{constraint2}\\
                    & F_{k,a}(t) \in \mathbb{Z}, \quad \forall k, \forall a, \forall t,\label{constraint3}\\
                    & 2 \leq F_{k,a}(t) \leq \xi \Delta, \quad \forall k, \forall a, \forall t,\label{constraint4}\\
                    & \mathrm{QoE}_k(t) \geq \mathrm{QoE}_{\text{th}}, \quad \forall t, \label{constraint5}\\
                    & \mathrm{hfQoE}(t) \geq \mathrm{hfQoE}_{\text{th}}, \quad \forall t.\label{constraint6}
\end{align}
\end{subequations}
Constraint \eqref{constraint1} stipulates the overall allocated bandwidth must remain below the threshold $b_{\text{max}}$. Meanwhile, the constraint \eqref{constraint2} establishes that the CPU frequency of the system for all users should not exceed the total frequency capacity of the server. The requirements outlined in constraints \eqref{constraint3} and \eqref{constraint4} dictate that the keyframes corresponding to various levels of user attention must be expressed as integers, ensuring compliance with the demands for communication and motion reconstruction. Furthermore, constraints \eqref{constraint5} and \eqref{constraint6} are implemented to guarantee $\mathrm{QoE}_k(t)$ and $\mathrm{hfQoE}(t)$ remain above the specified threshold.

\section{Causal-Aware Reinforcement Learning}
\label{algorithm}
In multi-user VR interactions, resource allocation poses a challenge that requires optimizing various interdependent factors, such as bandwidth, CPU frequency, and keyframe ratio, which include both continuous and discrete variables. This complexity makes traditional optimization techniques resource-intensive, particularly in real-time scenarios involving multiple participants. Furthermore, the allocated resources and the virtual environments heavily determine the perceived experience of users. Understanding this causal relationship is crucial as it can significantly facilitate the RL training process. As a result, this section provides a detailed explanation of the proposed PS-CDDPG algorithm to address problem ($\mathcal{P}$). We introduce partial state causal influence detection and noise-based active exploration strategies as general enhancements of the original framework, aiming to improve the training efficiency utilizing causal information.
\vspace{-.2cm}

\subsection{Problem Transformation}
RL-based algorithms are designed to make long-term decisions for MDP. They can continuously adjust resource allocation strategies based on current system conditions, enabling more flexible and context-sensitive decision-making. The definitions of the state space, action space, and reward function for the proposed system model within the RL framework are outlined as follows:
\begin{itemize}
    \item
     \textbf{State Space:} At time slot $t$, the state $S(t)$ is defined as:
     \begin{align}
        S(t)=&\{G_k(t),D_k(t-1), \mathrm{QoE}_k(t-1), \nonumber\\ 
        &\mathrm{avgQoE}_k(t), H(t), L(t)\}, \forall k \in \mathcal{K},         
     \end{align}
     where $G_k(t)$ indicates the attention data obtained from the virtual scene, $D_k(t-1)$ and $\mathrm{QoE}_k(t-1)$ are the delay and QoE value determined by the previous action. The other variables are used to compute $\mathrm{hfQoE}(t)$, which in turn enables effective long-term decision-making. In addition, we use $S^{\prime}(t)$ to denote the next state $S(t+1)$ for simplification.
    \item 
    \textbf{Action Space:} In accordance with the proposed optimization problem, the decisions involve the allocation of $\textit{\textbf{B}}$, $\textit{\textbf{f}}$, and $\textit{\textbf{F}}$. Consequently, the action $A(t)$ can be expressed as:
     \begin{equation}
        A(t)=\{b_k(t), f_k^e(t), F_{k,a}(t)\}, \forall k \in \mathcal{K},\forall a \in \alpha.
     \end{equation}      
    \item 
    \textbf{Reward:} The optimization objective consists of two aspects: first, to optimize the QoE within the constraints defined by \eqref{constraint1}-\eqref{constraint4}; and second, to ensure compliance with the thresholds established by \eqref{constraint5}-\eqref{constraint6}. The reward function at time slot $t$ can be expressed as follows:
     \begin{subequations}
     \begin{align}
        R(t)=&\mathcal{R}\left( S(t), A(t) \right)\\
            =&\sum_{k=1}^K \mathrm{QoE}_k(t)-\omega_1 \sum_{k=1}^K q_k^{\mathrm{QoE}}-\omega_2 q^{\mathrm{hfQoE}},
     \end{align}
     \label{reward}
     \end{subequations} 
    where $\omega_1$ and $\omega_2$ represent penalty coefficients, and $q_k^{\mathrm{QoE}}$ and $q^{\mathrm{hfQoE}}$ are binary variables indicating whether the given thresholds are satisfied (0) or violated (1).
\end{itemize}
\vspace{-.2cm}

\subsection{Causal Influence Detection}
For any given state $S(t) = s$ and action $A(t) = a$, causal influence detection differs from RL in that it emphasizes whether the current state is significantly influenced by the action taken, rather than solely on optimizing the actions. Through the quantity that measures whether the expected next state $S^{\prime}(t)$ is more controlled by the agent or the environment, we can guide the RL model to select better actions during exploration. Therefore, the CAI score is proposed based on conditional mutual information (CMI) which is zero for independence \cite{CRL_0}:
\begin{align}
    C(s,a,t) & = \frac{1}{|S^{\prime}|}\sum_{j=1}^{|S^{\prime}|}I(S^{\prime}_j(t);A(t)|S(t)=s,A(t)=a)\nonumber\\
    & = \frac{1}{|S^{\prime}|}\sum_{j=1}^{|S^{\prime}|}\text{D}_{\text{KL}}\left(p(s^{\prime}_j|s,a)\mid \mid p(s^{\prime}_j|s)\right),
\end{align}
where $j$ denotes the $j^{th}$ variable in next state, and $I(S^{\prime}_j;A|S=s)$ is the CMI between actions and next state, which can be calculated via the expectation of Kullback-Leibler (KL) divergence \cite{information_theory}. For the given state $s$, $p(s^{\prime}_j|s,a)$ represents the transition distribution of the next state when action $a$ is taken, while $p(s^{\prime}_j|s)$, also known as the transition marginal, denotes the expected distribution on all possible actions. When the CAI score is high, indicating that the KL divergence between $p(s^{\prime}_j|s,a)$ and $p(s^{\prime}_j|s)$ is significant, the action has a greater influence on the state $s$. Conversely, if the CAI score is approximately zero, the two distributions are similar, suggesting that the agent cannot control the evolution of the current state by taking action $a$. To calculate the KL divergence, we adopt the simplifying assumption that the transition distribution $p(s^{\prime}_j|s,a)$ is normally distributed given the state and action.

In the analysis of MDP, the nature of action variables plays a crucial role in determining the transition dynamics between states. For discrete problems, the calculation of $p(s^{\prime}_j|s)$ and the expectation of the KL divergence can be performed considering all possible actions and measuring the corresponding next state. However, in cases with continuous action variables, there are infinitely many actions within the action space. As a result, $p(s^{\prime}_j|s)$ must be computed through integration over the action space. Assuming that the action $a$ is governed by the policy $\pi(a|s)$, the formula for the transition marginal can be expressed as follows:
\begin{equation}
    p(s^{\prime}_j|s) = \int p(s^{\prime}_j|s,a)\pi(a|s)da.
\end{equation}

To compute this term and simplify the calculation, Monte-Carlo approximation method is applied to derive the outer expectation and the transition marginal by sampling $\mathcal{N}$ actions $\{ a^{(1)},a^{(2)},\dots,a^{(\mathcal{N})} \}$. Therefore, $C(s,a,t)$ can be derived with KL divergence, $\text{D}_{\text{KL}}$:
\begin{align}
    C(s,a,t) =
    \frac{1}{|S^{\prime}|}\sum_{j=1}^{|S^{\prime}|}\text{D}_{\text{KL}}(p(s^{\prime}_j|s,a)\mid \mid \frac{1}{\mathcal{N}}\sum_{n=1}^\mathcal{N} p(s^{\prime}_j|s,a^{(n)})).
    \label{CAI}
\end{align}

In addition, calculating the transition distribution for the high dimension scenario utilizing traditional methods is known to have limited performance \cite{kernel_method}\cite{kernel_method_2}. To address this limitation, a DNN is utilized as the inference model to approximate the probabilistic density function with high efficiency. With the normality assumption, the next state is parameterized as $S^{\prime}(t)\sim \mathcal{N}(\mu_\theta\left( s,a \right), \sigma_\theta^2\left( s,a \right))$, where $\theta$ is the network parameter. The inference model takes the states and actions as inputs and outputs the mean $\mu_\theta\left( s,a \right)$ and variance $\sigma_\theta^2\left( s,a \right)$ of each state variable in the next state. To train the inference model, $\theta$ can be updated by minimizing the log-likelihood of the sampled data $\mathcal{D^{C}}= \{\left( s^{(i)},a^{(i)},s^{\prime(i)}\right)\}^{|\mathcal{D^{C}}|}_{i=1}$ of the MDP process:
\begin{align}
    \theta^* = &\mathop{\text{argmin}}_{\theta} \frac{1}{|\mathcal{D^{C}}|}\sum_{i=1}^{|\mathcal{D^{C}}|}\frac{\left(s^{\prime(i)}-\mu_\theta\left(s^{(i)},a^{(i)}\right)\right)^2}{2\sigma_\theta^2\left(s^{(i)},a^{(i)}\right)}+\nonumber\\
    &\qquad\qquad\qquad\qquad\qquad\frac{1}{2}\log\sigma_\theta^2\left(s^{(i)},a^{(i)}\right).
    \label{log_loss}
\end{align}

Therefore, the inference model can be trained through iterations, and for the $j^{th}$ state variable, the corresponding CAI score $C_j(s)$ can be obtained. We denote the total number of training epochs as $\mathcal{I}_{max}$. The pseudocode of causal influence detection is given in \textbf{Algorithm \ref{Causal_Influence_Detection_algorithm}}.


\begin{algorithm}[htbp]
\caption{Causal Influence Detection} %

{\bf Input:} State $S(t) = s$, action $A(t) = a$, and replay memory\\
{\bf Output:} CAI score $C(s,a,t)$
\begin{algorithmic}[1]
\State \textbf{Initialization:} Network of inference model $f(S,A|\theta)$.

\State \textbf{\% Model Training}
\For {epochs $\mathcal{I}$ = 1 to $\mathcal{I}_{max}$}
    \State Select batch of data $\mathcal{D^{C}}= \{\left( s^{(i)},a^{(i)},s^{\prime(i)}\right)\}^{|\mathcal{D^{C}}|}_{i=1}$
    \For {data \textit{i}  = 1 to  $|\mathcal{D^{C}}|$}
        \State Obtain $\mu_\theta\left(s^{(i)},a^{(i)}\right)$ and $\sigma_\theta^2\left(s^{(i)},a^{(i)}\right)$
    \EndFor
    \State Update $f(S,A|\theta)$ utilize (\ref{log_loss})
\EndFor

\State \textbf{\% CAI score calculation}
\State Input $S(t) = s$ and $A(t) = a$ to  $f(S,A|\theta)$
\State Obtain $\mu_\theta\left(s,a\right)$ and $\sigma_\theta^2\left(s,a\right)$
\State Derive distributions $\{ p(s_j^{\prime}|s,a) \}_{j=1}^{|S^{\prime}|}$
\State Sample actions $\{ a^{(1)},a^{(2)},\dots,a^{(\mathcal{N})} \}$
\For{\textit{n} = 1 to  $\mathcal{N}$}
    \State Obtain $\mu_\theta\left(s,a^{(n)}\right)$ and $\sigma_\theta^2\left(s,a^{(n)}\right)$
    \State Derive distributions $\{ p(s_j^{\prime}|s,a^{(n)}) \}_{j=1}^{|S^{\prime}|}$
\EndFor
\State Calculate CAI score given by (\ref{CAI})

\end{algorithmic}
\label{Causal_Influence_Detection_algorithm}
\end{algorithm}
    \vspace{-.4cm}

\subsection{PS-CDDPG Algorithm}
As an actor-critic RL algorithm, DDPG is particularly effective in managing problems with continuous action spaces. In contrast to traditional deep Q-learning, which is designed for discrete action spaces, DDPG utilizes two deep neural networks: an actor network and a critic network, to make decisions. This architecture enables DDPG to optimize high-dimensional and continuous variables, making it well-suited for the proposed problem. Specifically, the actor network $\mathcal{X}(S|\vartheta_\mathcal{X})$ takes the state as input and approximates the policy function to output actions. The critic network $\mathcal{Q}(S, A|\vartheta_\mathcal{Q})$ computes the Q-values using the given state and action. To enhance the model's convergence and long-term reward potential, two target networks, $\mathcal{X}^-(S|\vartheta_\mathcal{X}^-)$ and $\mathcal{Q}^-(S, A|\vartheta_\mathcal{Q}^-)$, are employed during the training process. These target networks share the same architecture as the main networks but have different weights, with the parameters $\vartheta_\mathcal{X}^-$ and $\vartheta_\mathcal{Q}^-$ being updated using the soft update method. The pseudocode of the proposed PS-CDDPG framework is given in \textbf{Algorithm \ref{PS-CDDPG_algorithm}}.

\begin{algorithm}[htbp]
\caption{PS-CDDPG Algorithm} %

{\bf Input:} Human motion sequences of multi-user scenario\\
{\bf Output:} Resource allocation decisions
\begin{algorithmic}[1]
\State \textbf{Initialization:} Actor network $\mathcal{X}(S|\vartheta_\mathcal{X})$, critic network $\mathcal{Q}(S, A|\vartheta_\mathcal{Q})$, target actor networks $\mathcal{X}^-$ with parameter $\vartheta_\mathcal{X}^- \leftarrow \vartheta_\mathcal{X}$, target critic networks $\mathcal{Q}^-$ with parameter $\vartheta_\mathcal{Q}^- \leftarrow \vartheta_\mathcal{Q}$, inference model $f(S,A|\theta)$, and empty replay memory.

\State \textbf{\% Model Training}
\For {epochs $\mathcal{I}$ = 1 to $\mathcal{I}_{max}$}
    \State Initialize action-relevant state variables $S^{(2)}(0)$
    \For{\textit{t} = 1 to \textit{T}}
        \State Observe current state $S(t)=\{S^{(1)}(t), S^{(2)}(t)\}$
        \State Obtain action $A(t)=a$ from $\mathcal{X}(S(t)|\vartheta_\mathcal{X})$
        \State \textbf{\% Noise-based active exploration}
        \State Generate random number $\epsilon^*$
        \If{ $\epsilon^*\leq \epsilon$ }
            \State Generate a batch of noise $\{ \eta^{(1)},\eta^{(2)},\dots,\eta^{(\mathcal{N})} \}$
            \State Obtain candidate actions $\{ a+\eta^{(i)} \}_{i=1}^\mathcal{N}$
            \For {\textit{i} = 1 to $\mathcal{N}$}
                \State Derive distributions $\{ p(s_j^{\prime}|s,a+\eta^{(i)}) \}_{j=1}^{|S^{(2)}|}$ 
                \State Calculate CAI score $C(s,a+\eta^{(i)},t)$
            \EndFor
            \State Calculate priority weights according to (\ref{explore_weight})
            \State Select action $a^*$ as exploration result
        \EndIf
        \State \textbf{\% Model Training}
        \State Calculate reward $R(t)$ using (\ref{reward})
        \State Observe $S^{(1)}(t+1)$ and calculate $S^{(2)}(t+1)$
        \State Obtain $S(t+1)=\{S^{(1)}(t+1), S^{(2)}(t+1)\}$
        \State Store $\{S(t),A(t),R(t),S(t+1)\}$ in replay memory
        \State Select data $\mathcal{D^R}= \{\left( s^{(i)},a^{(i)},r^{(i)},s^{\prime(i)}\right)\}^{|\mathcal{D^{R}}|}_{i=1}$
        \State Update RL agent by (\ref{loss})
        \State Update inference model by \textbf{Algorithm \ref{Causal_Influence_Detection_algorithm}}
    \EndFor
\EndFor
\State \textbf{\% Decision making}
\State Initialize action-relevant state variables $S^{(2)}(0)$
\For{\textit{t} = 1 to $\infty$}
     \State Observe current state $S(t)=\{S^{(1)}(t), S^{(2)}(t)\}$
     \State Obtain the action $A(t)$ from actor network $\mathcal{X}(S(t)|\vartheta_\mathcal{X})$
\EndFor

\end{algorithmic}
\label{PS-CDDPG_algorithm}

\end{algorithm}

\begin{figure}[!t]
  \centering
\includegraphics[width=\linewidth]{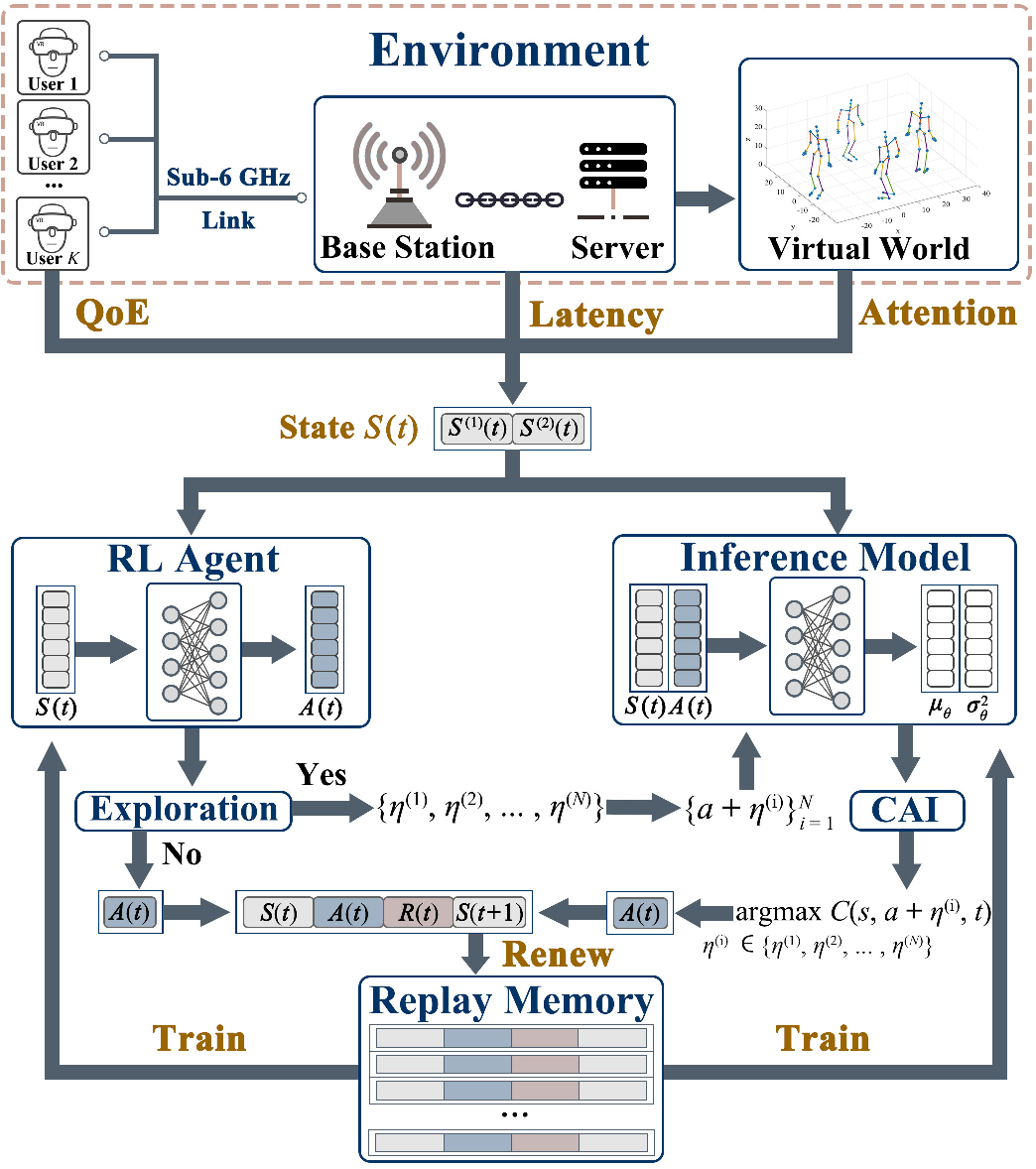}
  \caption{Framework of the proposed PS-CDDPG algorithm.}
  \label{Algorithm_fig}
      \vspace{-.6cm}
\end{figure}

Fig. \ref{Algorithm_fig} illustrates the framework of the PS-CDDPG algorithm and how CI can empower the training process with the assistance of CAI score. In real-world applications, the service can continue indefinitely, resulting in an infinite number of time slots. This scenario aligns with the formulation of problem ($\mathcal{P}$). However, during the training phase, the data sequence is finite. Therefore, we denote the total number of time slots in each data sequence as $T$. For time slot $t$, the state $S(t)$ is obtained from the environment, which is inputted to the RL agent to generate action $A(t)$, reward $R(t)$. Then the CAI score $C(s,a,t)$ can be determined by entering $S(t)$ and $A(t)$ into the inference model to generate the normal distribution parameters of $S^{\prime}(t)$. The replay memory stores the experience tuple $\{S(t), A(t), R(t), S^{\prime}(t)\}$ as a single training data point. Once the buffer fills to its capacity $M$, it eliminates older entries to make room for new ones. A random batch $\mathcal{D^R}= \{\left( s^{(i)},a^{(i)},r^{(i)},s^{\prime(i)}\right)\}^{|\mathcal{D^{R}}|}_{i=1}$ is then sampled from the buffer to train the inference model and the RL agent. The target network is used to calculate the expected $Q$-value for the agent:
\begin{equation}
    \mathcal{Q}^{(i)}_{targ}=r^{(i)}+\gamma\mathcal{Q}^-(s^{\prime(i)}, \mathcal{X}^-(s^{\prime(i)}|\vartheta_\mathcal{X}^-)|\vartheta_\mathcal{Q}^-),
\end{equation}
where $\gamma$ denotes the discount factor. The loss function for updating the critic network is defined as:

\begin{equation}
    L(\vartheta_\mathcal{Q})=\frac{1}{|\mathcal{\mathcal{D^R}}|}\sum_{i=1}^{|\mathcal{\mathcal{D^R}}|}\left( \mathcal{Q}^{(i)}_{targ}- \mathcal{Q}(s^{(i)}, a^{(i)}|\vartheta_\mathcal{Q})\right),
    \label{loss}
\end{equation}
where $|\mathcal{\mathcal{D^R}}|$ represents the batch size. The weight parameters of the critic network are effectively updated by minimizing $L(\vartheta_\mathcal{Q})$. Afterward, the actor network is trained to maximize the $Q$-value of the output actions using the policy gradient method.

For the inference model, the previous method attempts to measure the control of actions over all the variables in the next state. However, not all state variables can be influenced by the actions during application; for example, background noise, network fluctuations, and users' operations. For each time slot, the data is directly obtained from the environment and is important for decision-making, but it is not relevant to the previous action. In the proposed problem, the vision attention data $G_k(t)$ is irrelevant to the action because the attention target of each user cannot be altered by the allocated resources. Theoretically, regardless of the action taken, these action-irrelevant state variables will exhibit the same distribution in the next state, which will not affect the CAI score since the KL divergence is 0. However, the distribution is generated by the neural network, which cannot guarantee the same distribution for two different inputs during training iterations. This discrepancy can lead to additional loss during backpropagation and decrease inference accuracy. Therefore, inspired by the work in \cite{division}, we design the partial state framework of causal influence detection that divides the state $S(t)$ into action-irrelevant state variables $S^{(1)}(t)$ and action-relevant state variables $S^{(2)}(t)$. Both sets will be used as input, but only $S^{(2)}(t)$ will be considered for the calculation of the CAI score. Given the information of VR interaction system model, we can derive that $S^{(1)}(t) = \{G_k(t)\}$, which contains the attention data of users. In addition, $S^{(2)}(t)$ is composed of other variables included in $S(t)$. Notably, these variables represent the communication data that depend on user actions.

In addition, exploitation and exploration are essential ideas for RL training. For every step, if the agent chooses exploitation, it will take the action with the highest expected reward. If the agent chooses exploration, a random action will be taken to facilitate the agent to discover new states and actions in the environment. Random actions will add noise to the training data, which increases the total number of training iterations but also improves the accuracy. For the original RL algorithms, the $\epsilon$-greedy and noise-based methods are the basic exploration approaches. $\epsilon$ indicate the percentage of exploration, which is the key idea of $\epsilon$-greedy scheme. The work\cite{CRL_0} proposed the active exploration method based on causal influence detection and $\epsilon$-greedy method, which generate a batch of random actions for state $S(t)$ and select one according to the CAI value $C(s,a,t)$ as exploration. However, noise-based exploration is the main approach for continuous problems, which generate a random vector according to the given variance as noise and add it into the action given by agent to produce a new action. When an action space is complex, randomly generated actions are mostly meaningless actions with poor performance. Even if we can select the best action from the randomly generated actions during active exploration, the performance is still limited, and even affect the convergence. As a consequence, we exploit the above approaches to propose the noise-based active exploration method, which generate a batch of noise $\{ \eta^{(1)},\eta^{(2)},\dots,\eta^{(N)} \}$first and use the noise to generate candidate actions  $\{ a+\eta^{(i)} \}_{i=1}^N$: 
\begin{equation}
    a^* = a+\mathop{\text{argmax}}_{\eta^{(i)} \in \{ \eta^{(1)},\eta^{(2)},\dots,\eta^{(N)} \}} C(s,a+\eta^{(i)},t),
    \label{exploration}
\end{equation}
where $\eta^{(i)}\sim \mathcal{N}(0, \sigma_\eta^2)$ and $\sigma_\eta^2$ is the variance to determine the range of the noise. Furthermore, in addition to selecting the action with the highest CAI value, the original work also adopts a weight-based method for action selection. Specifically, the probability of choosing the $i^{th}$ action can be derived from:
\begin{equation}
    p^{(i)}=\frac{N-\mathcal{R}^{(i)}}{\sum_j\mathcal{R}^{(j)}}.
    \label{explore_weight}
\end{equation}
In this equation, $\mathcal{R}^{(i)}$ denotes the rank of the $i^{th}$ action when sorting all $N$ actions according to their CAI scores ${C(s,a+\eta^{(i)},t)}_{i=1}^N$. In particular, for actions with a higher influence, the rank will be smaller, which yields a higher priority for selection during the exploration phase.

To ensure the DDPG algorithm conforms to the MIP problem formulation, constraints are introduced to the output actions. The output layer uses the Sigmoid function as its activation function to restrict each action component to the interval (0, 1). Assuming the output action is given by $\hat{A}(t)=\{\hat{b}_k(t), \hat{f}_k^e(t), \hat{F}_{k,a}(t)\}$, these values are normalized to indicate the ratio of the total resources:
\begin{align}
\left\{
	\begin{aligned}
	&b_k(t)=\frac{1}{\sum_{i=1}^K\hat{b}_i(t)}\hat{b}_k(t), \\
	&f^e_k(t)=\frac{1}{\sum_{i=1}^K\hat{f}^e_i(t)}\hat{f}^e_k(t).
	\end{aligned}
\right.
\end{align}
This guarantees that the total allocation ratios for bandwidth and frequency resources sum to 1.  The keyframe ratio $F_{k,a}(t)$ is derived by discretizing $\hat{F}_{k,a}(t)$ in accordance with the frames per second (FPS) $\xi$:
\begin{equation}
    F_{k,a}(t)=\frac{1}{\xi}\lceil \hat{F}_{k,a}(t) \xi \rceil,
\end{equation}
The ceiling function, denoted by $\lceil x \rceil$, returns the smallest integer greater than or equal to $x$. The resulting keyframe ratio $F_{k,a}(t)$ will produce an integer number of keyframes when multiplied by $\xi$, in contrast to $\hat{F}_{k,a}(t)\xi$, which can produce any real number.

\section{Evaluation}
\label{evaluation}
This section outlines our evaluation methodology and presents a performance comparison with several baseline approaches. We utilize the CMU Graphics Lab Motion Capture Database\footnote{http://mocap.cs.cmu.edu}, which is widely utilized in human motion research and provided in BVH format. This database includes 2,605 motion sequences across 6 categories and 23 subcategories, sourced from recordings of 111 human subjects.

In the RL agent, the actor and critic networks are designed with 4 hidden layers, each containing $n_{r}=32$ neurons. The causal inference model contains 3 hidden layers with $n_{c}=256$ neurons each. The experiments are done with PyTorch in Python 3.9.19. All training and evaluation experiments are performed on an NVIDIA GeForce RTX 4070 GPU and an Intel Core i7-13700KF CPU with 64GB of RAM. Table \ref{parameters} summarizes the key parameters used for environment simulation and model training.

\subsection{Parameter Settings}
For our evaluation setup, we simulate a VR environment with 5 users interacting within a virtual scene with maximum dimensions of $10m \times 10m$. The FPS is maintained as 30, with each time slot having a duration of $\Delta t = 1s$. To ensure consistency,  we set the total number of time slots for each data sequence as $T=100$, excluding any sequences with fewer than $3000$ frames. To generate multi-user interaction data, five sequences are randomly selected, rotated, and translated to create augmented motion data. As a result of this process, we obtained a total of $2500$ datasets, assigning $2000$ for training and 500 for testing. We convert all coordinates from Euler angles to global positions, while the vision direction of each user is obtained using the joint coordinates of the head and shoulders.

In order to assess distances in the physical world,  the server is centrally located at $P_0=[0m, 0m]$ within the environment,  while the five users are randomly placed at the following locations: $P_1=[150m, 250m]$, $P_2=[-120m, 300m]$, $P_3=[0m, -300m]$, $P_4=[-280m, -50m]$, and $P_5=[100m, -320m]$\footnote{Note that these distances represent the communication distance between each user and the access point, rather than the interaction distance among users in the VR environment.}. According to the model presented in earlier research \cite{sub6G_2}, we define the reference distance as $d_0 = 1m$. The transmitter power level, $P_k(t)$, is set to $20dBm$, and the indoor noise power is $-110dBm$. The path loss, as detailed in Eq. (\ref{pl}), is expressed by the following equation, including shadow fading with a standard deviation of $\sigma=0.9$dB:
\begin{equation}
    PL(d)[dB]=49.12+12.4log_{10}\frac{d}{d_0}+X_\sigma.
\end{equation}
      \vspace{-0.4cm}

\begin{table}[ht]

\renewcommand{\tablename}{TABLE}
\caption{Evaluation Parameters}   

\centering
\begin{tabular}{cc|cc}
\toprule
\multicolumn{4}{c}{\textbf{Environment Parameters}} \\
\midrule  
\textbf{Parameters} & \textbf{Values} & \textbf{Parameters} & \textbf{Values} \\
\midrule  
$b_{max}$ & 10MHz & $f_{max}$ & 10GHz  \\
K & 5 & $f^r_k(t)$ & [1.5-2.5]GHz  \\
$\xi$ & 30 & $\delta$ & 10kB \\
$T_{max}$ & 150ms & $\Delta t $ & 1s\\
$\omega$ & 3 & $c_e$ & 30 Cycles/Bit\\
$c_r^1$ & 50 Cycles/Bit & $c_r^2$ & 240 Cycles/Bit\\
$P_k(t)$ & 20dBm & $\sigma_k^2$ & -110dBm\\
$\mathrm{QoE}_{th}$ & 0.2 & $\mathrm{hfQoE}_{th}$ & 0.6\\
\midrule  
\midrule  
\multicolumn{4}{c}{\textbf{Algorithm Parameters}} \\
\midrule  
\textbf{Parameters} & \textbf{Values} & \textbf{Parameters} & \textbf{Values} \\
\midrule  
$\gamma$ & 0.99 & $\tau$ & 0.01 \\
$n_{r}$ & 32 & $n_{c}$ & 256 \\
$lr_{\mathcal{Q}}$ & $6 \times 10^{-7}$ &$lr_{\mathcal{X}}$ & $5 \times 10^{-5}$ \\
$lr_{\mathcal{C}}$ & $1 \times 10^{-4}$ &$\omega_1$, $\omega_2$ & 0.5, 0.5\\
$|\mathcal{\mathcal{D^R}}|$ & 64 & $|\mathcal{D^C}|$ & 128\\
$N$ & 64  & $M$ & 5000 \\
$\epsilon$ &  0.4 &$\sigma_\eta^2$ & 0.01 \\
$\mathcal{I}_{max}$ & 10000 &$T$ & 100 \\

\bottomrule  
\vspace{-.85cm}
\label{parameters}
\end{tabular}
\end{table}

\subsection{Benchmark Comparison}
Causal influence detection acts as an assistant in RL training, enhancing training performance by guiding exploration without altering the decision-making process of the actor. As a result, the primary improvement achieved is the acceleration of training efficiency. Moreover, the proposed approach integrates state division and noise-based active exploration, which requires an evaluation of the effectiveness of these enhanced modules. To this end, we selected the following baseline approaches to compare against the proposed method:
\begin{itemize}
    \item \textit{DDPG:} The original DDPG algorithm without causal influence detection. The exploration method is the traditional noise-based approach.
    \item \textit{CAI+DDPG:} The original CRL method based on DDPG without state division. The exploration method is the proposed noise-based active approach.
    \item \textit{PS-CDDPG:} The proposed causal influence detection-based DDPG with state division and noise-based active exploration.
\end{itemize}

\begin{figure}
    \centering
    \includegraphics[width=0.95\linewidth]{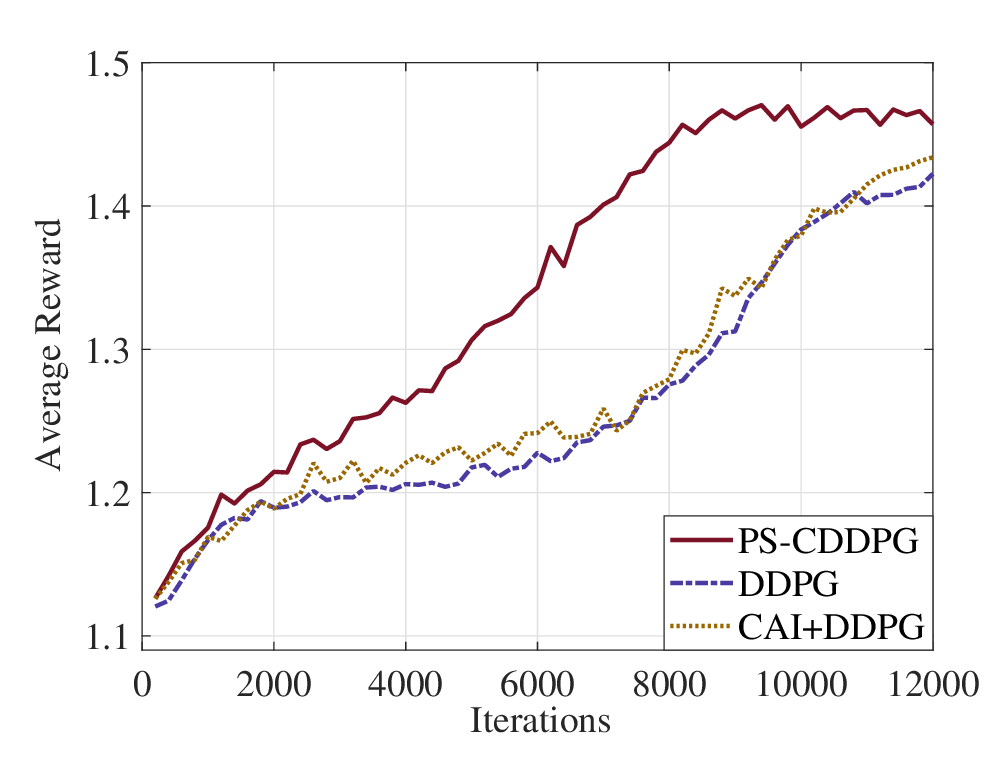}
    \caption{Average reward of different frameworks of CRL.}
    \label{baselines_reward}
      \vspace{-0.4cm}
\end{figure}

\begin{figure}
    \centering
    \includegraphics[width=0.95\linewidth]{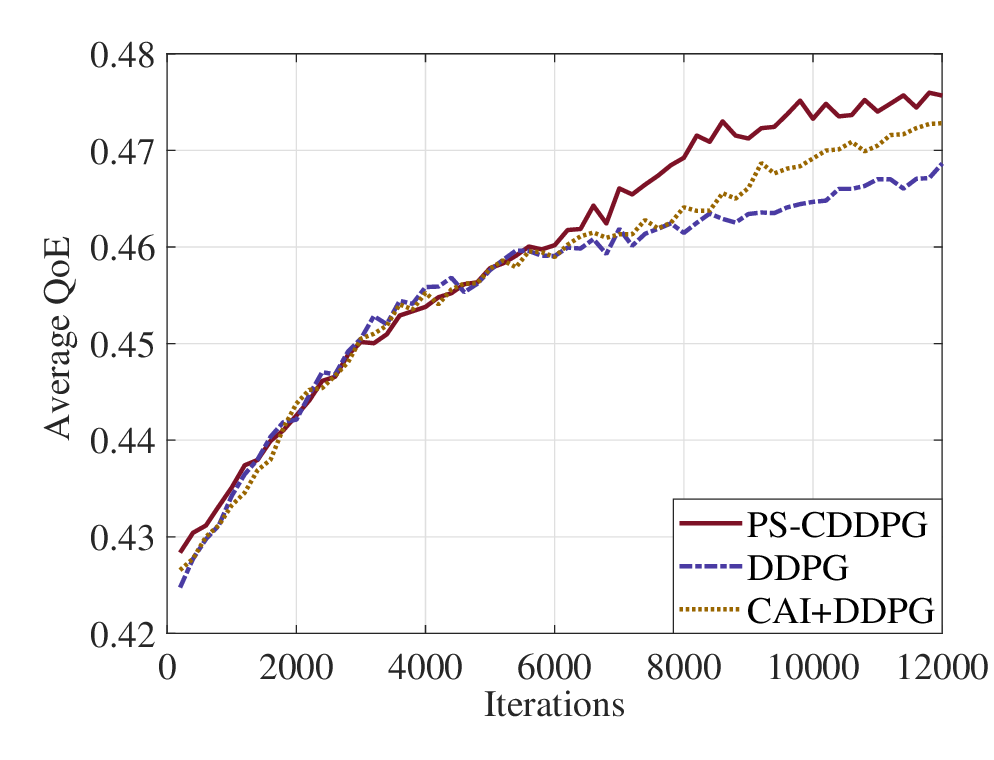}
    \caption{Average QoE of different frameworks of CRL.}
    \label{baselines_QoE}
      \vspace{-0.4cm}
\end{figure}

Fig. \ref{baselines_reward} and Fig. \ref{baselines_QoE} demonstrate the average reward and QoE across different frameworks. While it is important to consider the original causal influence detection-based method utilizing active exploration, its results were excluded due to the model's failure to converge. From the figures, the proposed method exhibits the best convergence performance and the highest accuracy throughout the training process. Specifically, it is observed that 30\% of the iterations can be eliminated while still achieving the best rewards, which exceed 1.45. Moreover, the results of QoE are influenced by the actions taken and the current state, particularly the user's attention. Causal influence detection investigates this kind of causal information and provides guidance throughout the exploration process. By strategically selecting actions that positively impact QoE, we can align the resulting QoE more closely with the reward requirements, thereby reducing randomness in the outcomes. Furthermore, simulation results indicate that the improvement in reward is more significant than in QoE. This observation can be attributed to the fact that the reward encompasses multiple measurement aspects, while the CAI score effectively maintains fairness in actions during exploration. 

Additionally, if all state variables are inferred within CDDPG, there is almost no improvement compared to DDPG. This shows that the inference model tries to approximate the distribution of action-irreverent state variables and become unfit. Therefore, the CAI score calculated on the basis of the CAI model will not be able to guide the selection of the action during exploration. The selected action during noise-based active exploration will be similar to a random action given by noise-based exploration, resulting in the performance similar to DDPG. The enhancement of state division and exploration approach can ensure the training accuracy of the inference model and improve the training efficiency. Although DDPG also has the possibility to achieve the same average reward if the iteration is extended, the training cost will be much higher and influence its effectiveness. Consequently, this structured approach not only enhances the overall performance of the proposed method but also underscores the importance of integrating causal analysis in the exploration framework.

\subsection{Sensitivity Analysis}

\begin{figure*}[!t]
  \centering
\includegraphics[width=\linewidth]{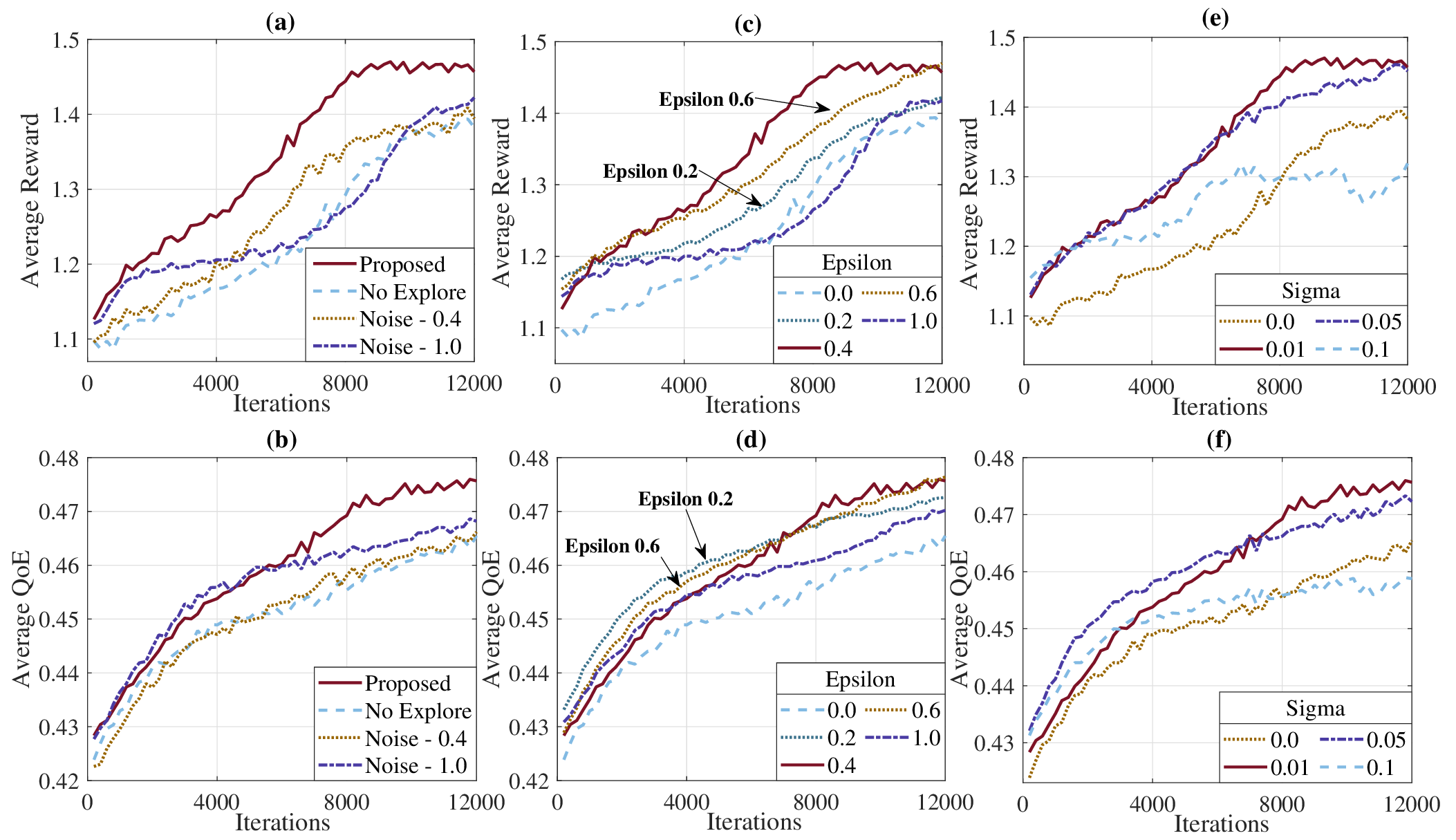}
  \caption{Comparison of convergence performance: (a) average reward of different exploration methods (b) average QoE of different exploration methods (c) average reward of different $\epsilon$ (d) average QoE of different $\epsilon$ (e) average reward of different $\sigma_\eta^2$ (f) average QoE of different $\sigma_\eta^2$.}
  \label{Convergence}
      \vspace{-0.8cm}
\end{figure*}

For the method based on $\epsilon$-greedy, $\epsilon$ is the hyperparameter to control the percentage of exploration, which is set to 0.4 in the proposed PS-CDDPG method. For the traditional noise-based algorithm, it is mostly not considered and is usually set to 1. To compare the influence of different exploration methods, we evaluate the performance of the proposed scheme, exploration with noise only, and without exploration. From Figs. \ref{Convergence} (a) and (b), it can be envisioned that when $\epsilon$ is small for the traditional noise-based exploration, there will be insufficient exploration actions and the improvement is limited, resulting in similar results without exploration. Also, when $\epsilon$ is set to one, which is commonly used for DDPG, the accuracy can be improved, but the training process will be less stable because of adding more noise to the actions. Our method increases exploration efficiency while avoiding the impact of noise on training stability. As a result, it demonstrates higher speed and stability compared to other methods, regardless of whether the commonly used setting ($\epsilon$ = 1) or the same setting ($\epsilon$ = 0.4) is applied.

The value of $\sigma_\eta^2$ also affects the exploration process, which determines the variance of the generated noise. To reveal the influence on the proposed noise-based active exploration scheme, we first fix the value of $\sigma_\eta^2$ at 0.01 and compare the convergence results as the value of $\epsilon$ ranges from 0 to 1. Then, we select $\epsilon$ to be 0.4 and compare the performance with different values of $\sigma_\eta^2$. Figs. \ref{Convergence} (c)-(f) further illustrate the impact of $\epsilon$ and $\sigma_\eta^2$ on the average return and QoE of the proposed method. Both parameters influence the proportion of noise during the exploration process. When the values of $\epsilon$ and $\sigma_\eta^2$ are small, the improvement is not significant, resembling the performance without exploration. However, if the hyperparameters are set too high, excessive noise can affect the accuracy of the inference model. When the CAI values are inaccurate, the model's performance tends to resemble that of exploration using only noise with the same parameter settings. Furthermore, when $\sigma_\eta^2$ is too large, overly random actions can even hinder the learning of the RL agent, thereby reducing the precision of decision-making.

\subsection{Adaptive Keyframe Extraction}


After systematically analyzing the convergence performance of the proposed PS-CDDPG algorithm, we select the most appropriate hyperparameter settings and record the converged model with the highest average reward to compare decision-making abilities. To evaluate the efficiency of the proposed approach, we compare it against the following baseline methods for multi-user interaction:
\begin{itemize}
    \item
    \textit{Original:} 
    All frames are transmitted without keyframe extraction or attention-based adaptation, with bandwidth uniformly allocated among users.
    \item
    \textit{Attention Only:} 
    The transmission includes only characters in the user’s FoV, based on an attention model, but with uniform bandwidth allocation.
    \item
    \textit{Keyframe Ratio of $33\% /50\% /66\%$:} Fixed The keyframe ratio as $33\% /50\% /66\%$ for different levels of attention with bandwidth and computational power uniformly allocated.
    \item
    \textit{Adaptive Keyframe:} The approach adaptively adjusts the keyframe ratio for different levels of attention.

\end{itemize}

\begin{figure}[htbp]
    \centering
    \includegraphics[width=0.95\linewidth]{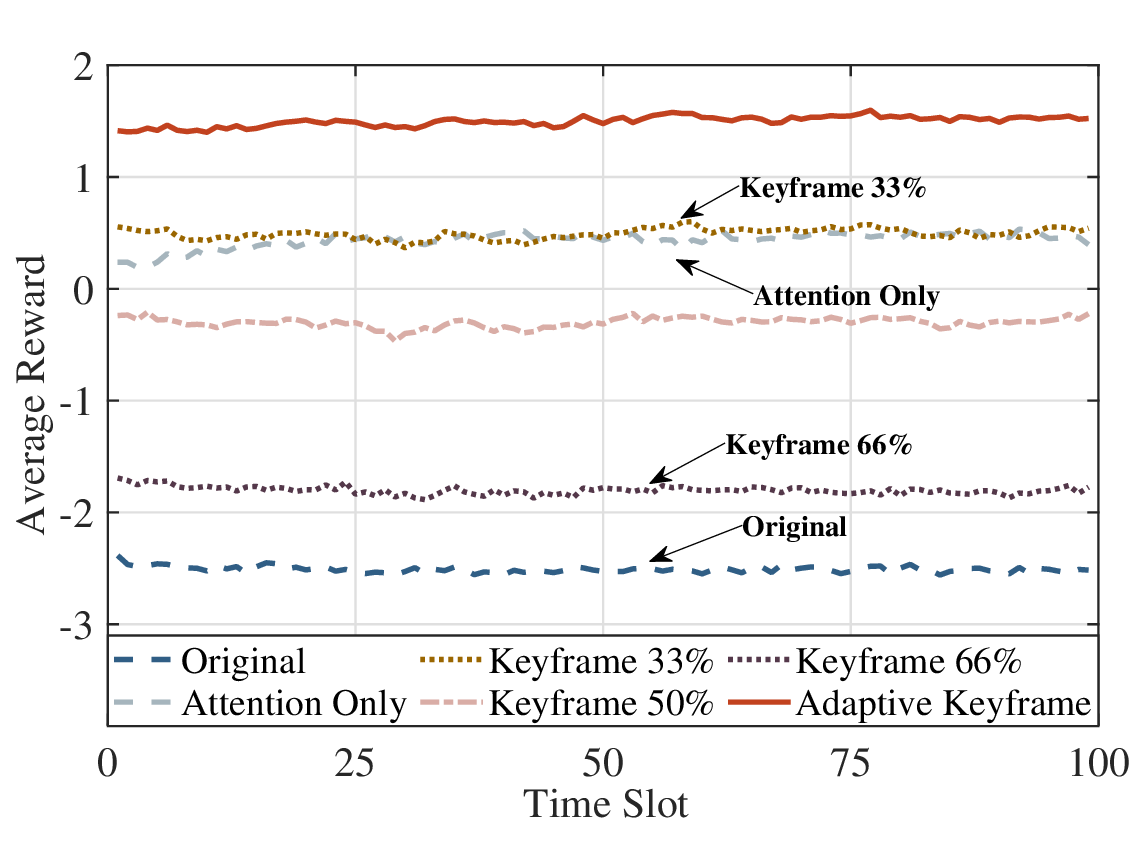}
    \caption{Comparison of the average reward across different time slots.}
    \label{reward_average}
    \vspace{-.4cm}
\end{figure}

Initially, we assess and compare the average reward across various schemes during different time slots.  As depicted in Fig. \ref{reward_average}, the average reward per time slot is derived from $500$ test data. During each time slot, we tested the reward from five users for each data sequence and calculated the average reward. It is clear that the adaptive keyframe technique provides the highest QoE performance while meeting the constraints specified in \eqref{constraint5} and \eqref{constraint6}. Despite the complexity of the environment and user actions, the average reward remained stable throughout the entire period.

\begin{figure}
    \centering
    \includegraphics[width=0.95\linewidth]{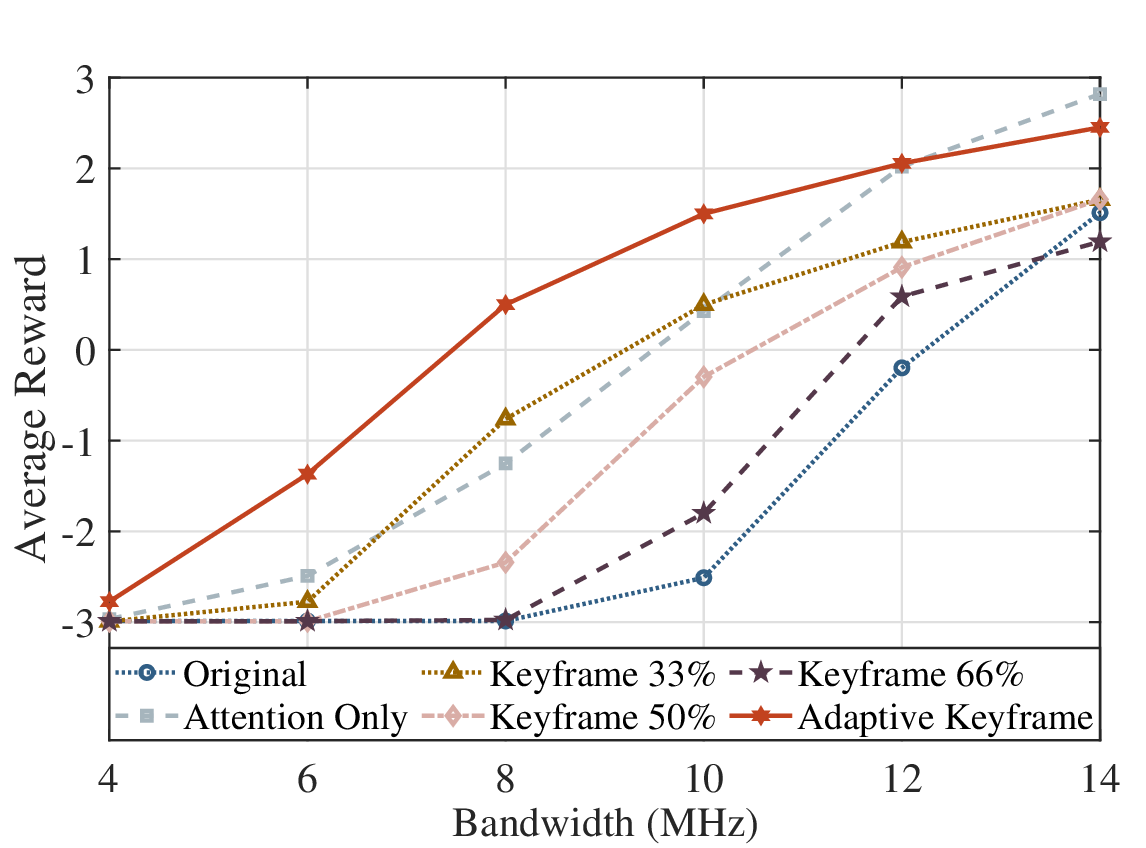}
    \caption{Comparison of the average reward across different bandwidth.}
    \label{reward_B}
    \vspace{-.4cm}
\end{figure}

Fig. \ref{reward_B} illustrates the impact of modifying the total bandwidth $b_{max}$. Although all schemes exhibit comparable performance in extremely high or low bandwidth conditions, our approach shows significant advances in scenarios with limited bandwidth. Once trained, the model can automatically adjust to different interactive scenarios and produce consistent decisions. By using keyframe-based and attention-driven approaches before transmission, the dependency on bandwidth can be significantly reduced.

\begin{figure}
    \centering
    \includegraphics[width=0.95\linewidth]{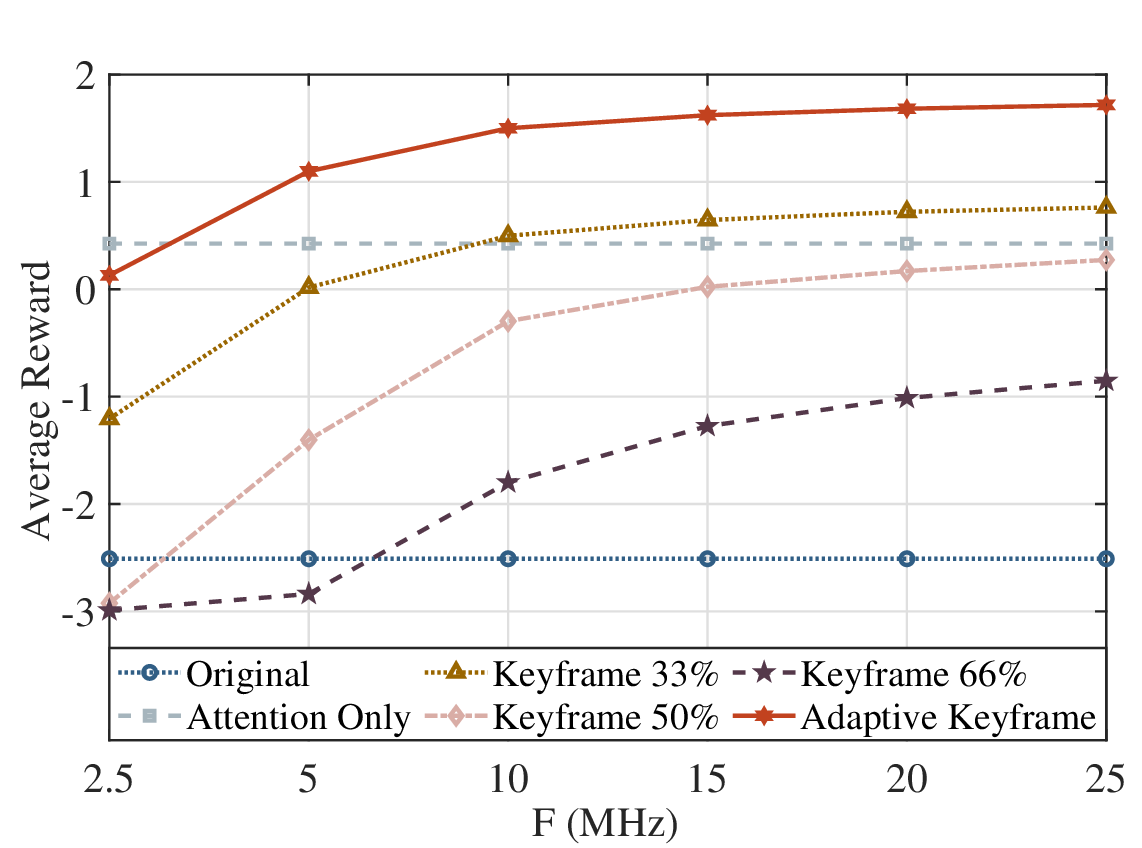}
    \caption{Average reward of Different CPU Frequencies.}
    \label{reward_F}
    \vspace{-.4cm}
\end{figure}

Additionally, we set the maximum bandwidth to 10MHz and explore the impact of different computational power $F$ on user experience. Fig. \ref{reward_F} illustrates the average reward of the baselines in response to changes in $F$. Since the original and the attention only method do not involve the keyframe extraction process, their performance remains unaffected by the CPU frequency. When $F$ is low, the time consumption for extracting the keyframes tends to be high, resulting in a high latency or a low keyframe ratio. As computational power increases, the improvement brought by keyframes becomes progressively apparent, leading our algorithm to exhibit the best performance.

\subsection{Impact of Different Parameters}

\begin{figure}[!t]
  \centering
        \includegraphics[width=0.95\linewidth]{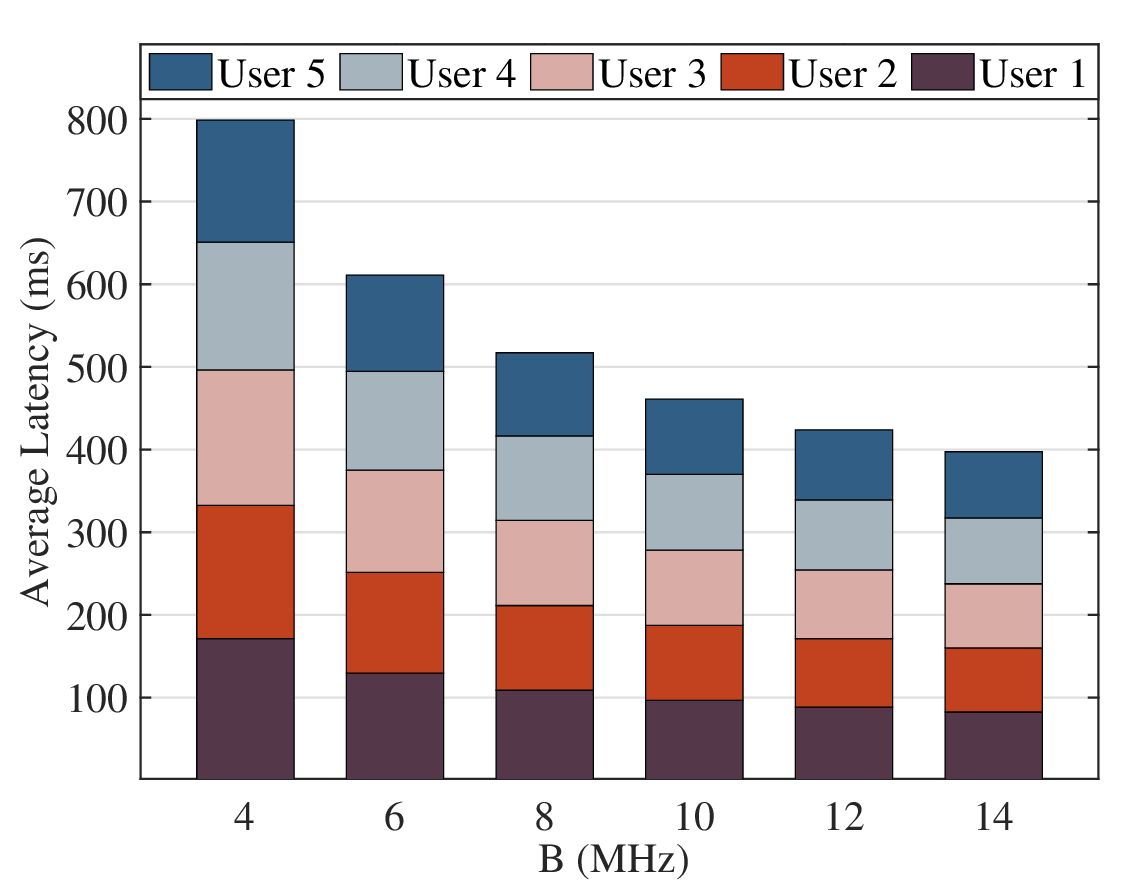}
        \caption{Average Latency of Different Bandwidth.}
        \label{latency_B}
        \vspace{-.4cm}
\end{figure}

To further evaluate the impact of different environmental parameters on the proposed method, we tested the latency and QoE performance of each user under different bandwidths and CPU frequencies. Fig. \ref{latency_B} further illustrates the variations in delay experienced by different users when bandwidth changes. As bandwidth increases, the average delay for each user decreases consistently. This demonstrates that our model can effectively balance delay reduction while enhancing rewards, thus adapting to varying bandwidth conditions.

\begin{figure}[!t]
  \centering
        \includegraphics[width=0.95\linewidth]{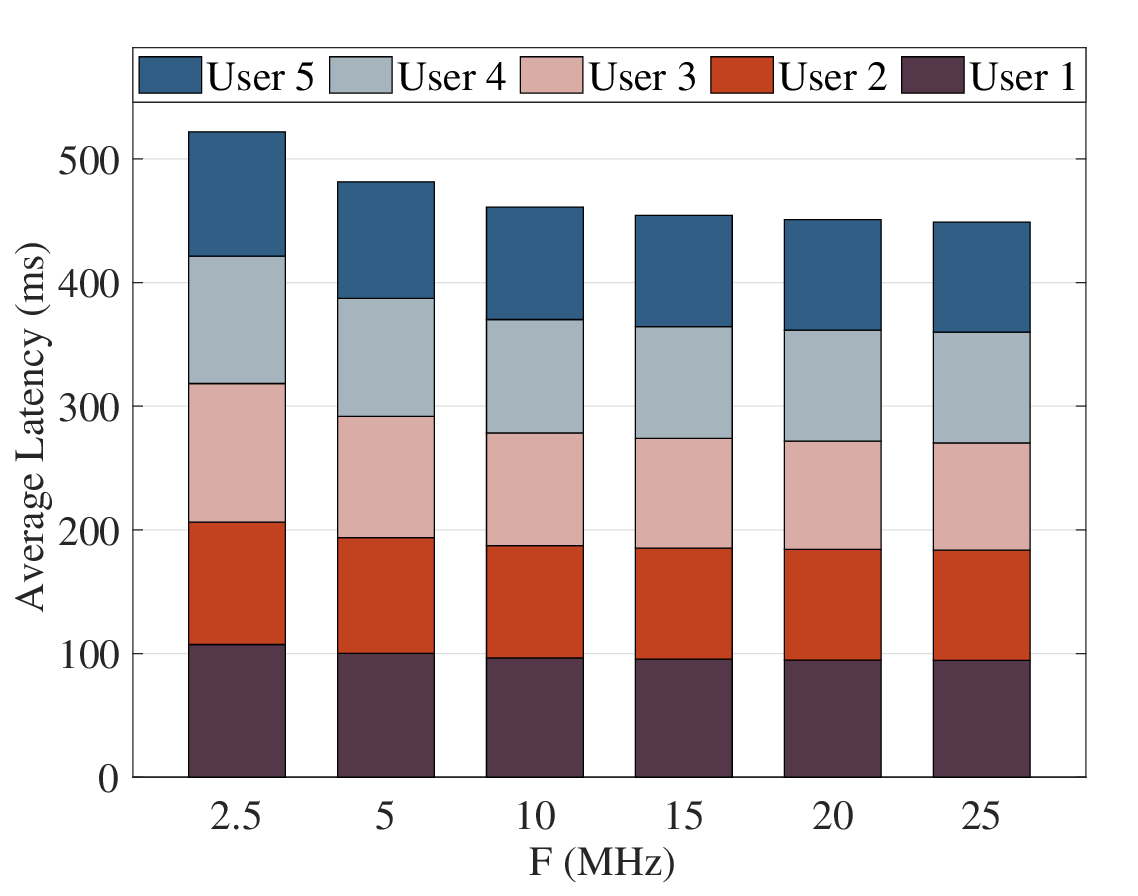}
        \caption{Average Latency of Different CPU Frequencies.}
        \label{latency_F}
            \vspace{-.4cm}
\end{figure}

Furthermore, Fig. \ref{latency_F} depicts the total latency for different users. Although the efficiency of keyframe extraction is low when computational power is insufficient, the differences in latency shown in the figure are not significant, which implies that the proposed model can selectively choose the ratio of keyframes based on computational resources, preventing excessive keyframe extraction from causing high latency.

\begin{figure}
    \centering
    \includegraphics[width=0.95\linewidth]{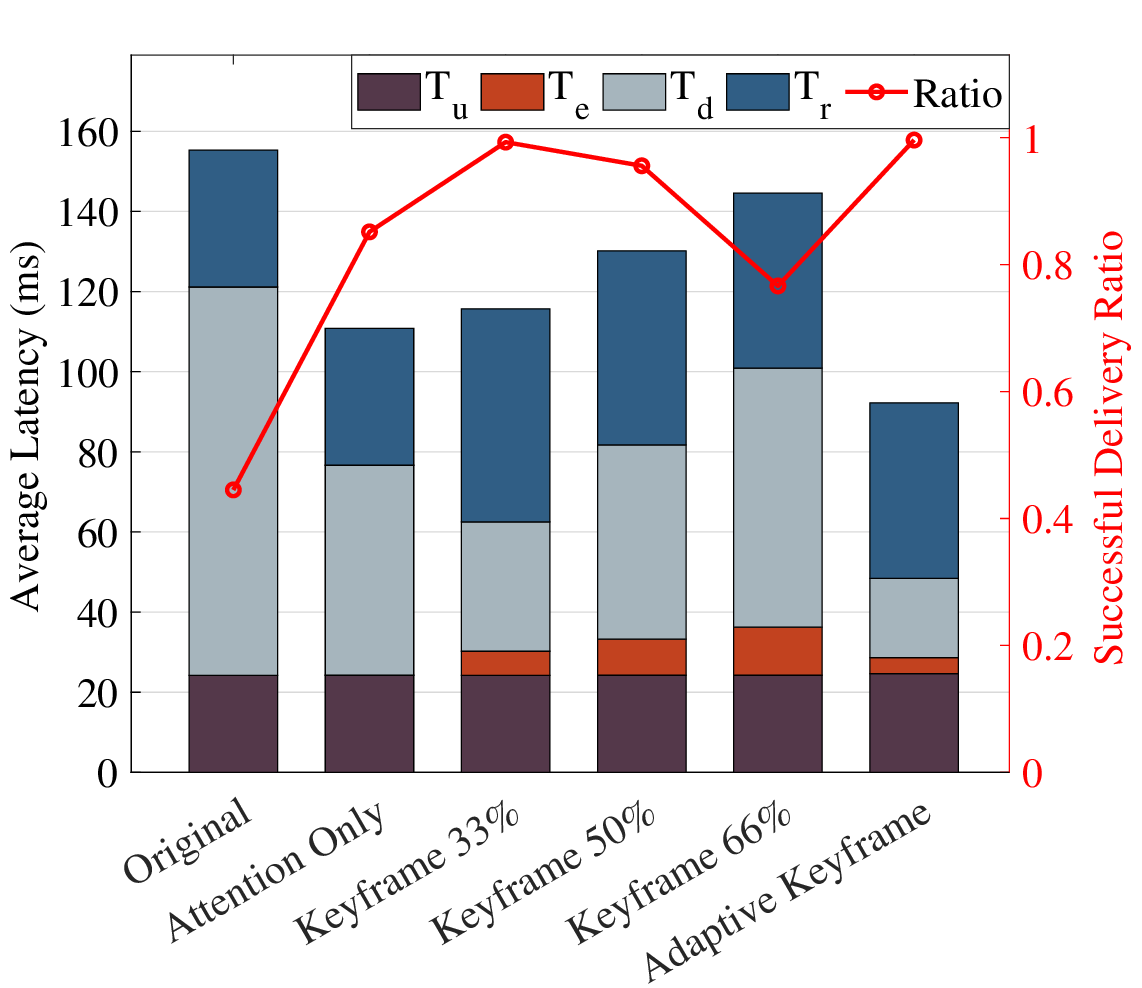}
    \caption{Comparison of different baselines in latency and successful delivery ratio.}
    \label{latency_baseline}
        \vspace{-.4cm}
\end{figure}

We then assess the average latency, QoE, delivery success rate, and fairness for different baselines.  As illustrated in Fig. \ref{latency_baseline}, the detailed time delays associated with different processes and the resulting successful delivery ratios are detailed. The selection of frames based on attention mechanisms substantially reduces download latency by sending only the necessary frames. Extracting keyframes increases computational effort but also reduces download latency, thereby improving the successful delivery ratio. Nonetheless, the efficacy of the keyframe-based method can vary considerably with different keyframe ratios. Our strategy effectively merges the advantages of attention mechanisms with keyframe transmission, improving QoE through adaptive keyframe ratio adjustments. Additionally, efficient allocation of bandwidth and CPU frequency among users leads to a reduction in the time required for keyframe extraction, denoted as $T_e$. Consequently, our method decreases average latency and raises the transmission success rate to above 99\%.

\begin{figure}
    \centering
    \includegraphics[width=0.95\linewidth]{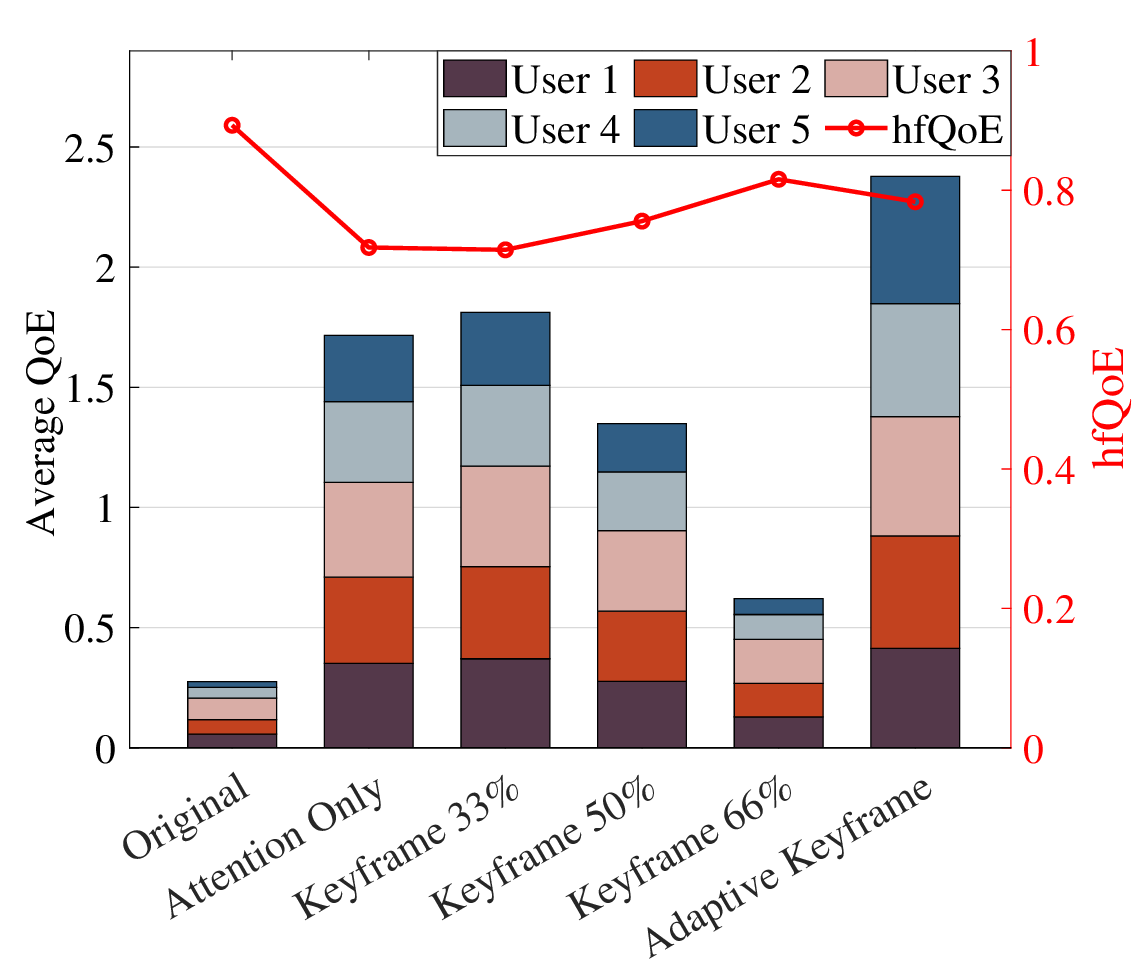}
    \caption{Comparison of different baselines in QoE and hfQoE.}
    \label{QoE_baseline}
        \vspace{-.4cm}
\end{figure}

Simply lowering latency does not inherently improve QoE; it is also important to take into account the negative effects of overly low keyframe ratios. As illustrated in Fig. \ref{QoE_baseline}, the proposed framework achieves both the highest average QoE and promising fairness. This advancement is attributed to effective data volume management during transmission and optimized resource allocation. The Original and Fixed Keyframe Ratio 66\% benchmarks exhibit high fairness largely due to their uniformly low QoE among all users, which limits variance. Conversely, our strategy yields a much higher QoE while maintaining fair performance among users.

\section{Conclusion and Future Work}
\label{conclusion}
In this paper, we proposed an innovative framework to improve the QoE in multi-user VR interactions with sub-6 GHz communications. Our approach incorporates attention-based strategy, keyframe extraction, and the Weber-Fechner Law, leading to the formulation of the QoE optimization challenge as a MIP problem. The PS-CDDPG algorithm is proposed by combining the DDPG method with causal influence detection for dynamic decision-making and efficient model training. The state division approach and noise-based active exploration are introduced to enhance the performance of the inference model. Experiments conducted with the CMU Motion Capture Database indicated that our framework enhances the convergence performance significantly, with more than 30\% iterations reduced to achieve the desired average reward. our approach also significantly reduces latency, ensures fairness, and enhances QoE, outperforming baseline approaches.

Moreover, this work also indicates that the existing causal influence detection algorithm has limitations when dealing with complex environments. Using only DNN as the inference model makes it difficult to accurately approximate the distribution of the next state variables, thereby affecting the accuracy of the inference model and the training of RL agent. Therefore, it becomes particularly important to eliminate the influence of action-irrelevant variables from the states and use noise to restrict the region of active exploration to improve the reliability of the inference model. In the future, we are planning to consider advanced models other than DNN as inference models to further enhance their effectiveness. 


\bibliographystyle{IEEEtran}
\bibliography{Reference}

\end{document}